\documentclass{article}
\usepackage{iclr2027_conference,times}

\usepackage{amsmath,amsfonts,bm}

\def\eqref#1{equation~\ref{#1}}

\def\1{\bm{1}}

\DeclareMathAlphabet{\mathsfit}{\encodingdefault}{\sfdefault}{m}{sl}
\SetMathAlphabet{\mathsfit}{bold}{\encodingdefault}{\sfdefault}{bx}{n}

\usepackage{hyperref}
\hypersetup{hidelinks}
\usepackage{url}
\usepackage{graphicx}
\graphicspath{{figs/}}
\usepackage{booktabs}
\usepackage{longtable}
\usepackage{amsmath,amssymb}
\usepackage{xcolor}
\usepackage{pifont}
\usepackage{tikz}
\usetikzlibrary{arrows.meta,positioning}
\usepackage{enumitem}
\definecolor{saRed}{RGB}{178,34,34}
\definecolor{saBlue}{RGB}{25,90,160}
\newcommand{\bad}[1]{\textcolor{saRed}{#1}}
\newcommand{\ok}[1]{\textcolor{saBlue}{#1}}
\newcommand{\yes}{\textcolor{saBlue}{\ding{51}}}
\def\xmark{{\color{black!42}\ding{55}}}
\usepackage{algorithm}
\usepackage{algorithmic}
\usepackage{float}
\usepackage{subcaption}
\usepackage{wrapfig}

\newcommand{\sysname}{\textsc{SelfEvoSkill}}

\newcommand{\passratefullvalue}{73.9}
\newcommand{\passratenoskillvalue}{40.9}
\newcommand{\passratecuratedvalue}{56.7}
\newcommand{\passratefull}{\passratefullvalue\%}

\newcommand{\deltavsnoskill}{+33.0}
\newcommand{\deltavscurated}{+17.2}
\newcommand{\numtasks}{89}
\newcommand{\numdomains}{8}

\newcommand{\TongyiLogo}{\AddToShipoutPictureFG*{\AtPageUpperLeft{\put(\LenToUnit{\dimexpr\paperwidth-5cm\relax},\LenToUnit{-3.5cm}){\includegraphics[width=1.5cm]{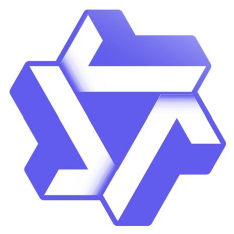}}}}}

\iclrfinalcopy

\begin{document}

\title{\sysname{}: Paired Execution Audits \\ for Skill Revision without \\External Outcome Supervision}

\author{
\textbf{Haowen Gao\textsuperscript{1,2},
Haoran Chen\textsuperscript{3}\thanks{Project leader.},
Can Wang\textsuperscript{3},
Shasha Guo\textsuperscript{1},
Liang Pang\textsuperscript{1}\thanks{Corresponding author.},} \\
\textbf{Zhaoyang Liu\textsuperscript{3},
Huawei Shen\textsuperscript{1},
Xueqi Cheng\textsuperscript{1}} \\
\\
\textsuperscript{1} State Key Laboratory of AI Safety, Institute of Computing Technology, CAS, Beijing, China \\
\textsuperscript{2} University of Chinese Academy of Sciences, Beijing, China \\
\textsuperscript{3} Tongyi Lab, Alibaba Group, Beijing, China \\
\\
{\small\texttt{gaohaowen23s@ict.ac.cn, pangliang@ict.ac.cn}}
}

\maketitle
\lhead{}
\TongyiLogo

\begin{abstract}
Agent skills provide LLM agents with reusable procedures, but a given skill may be incomplete, ineffective, or misleading for the current task. Existing methods for revising skills typically rely on external outcome supervision, such as task rewards, performance on a separate validation set, or feedback from deployment, to decide which updates to retain. However, this supervision may be costly or unavailable when adapting a skill to a new task. Using execution behavior as an alternative signal is challenging: a single run reflects the combined effects of the agent, task, and workspace, making the skill's contribution difficult to isolate. Comparing executions with and without the skill, while holding these factors fixed, provides evidence about how the skill changes behavior. We introduce \textbf{\sysname{}}, which links these behavioral differences to relevant skill passages and iteratively revises and reevaluates those passages through subsequent executions. Checks derived once from the task's explicit requirements prevent revisions from violating task constraints. Across \numtasks{} containerized tasks in \numdomains{} professional domains, under the same best-of-two evaluation budget, \sysname{} attains a \passratefull{} mean task reward, exceeding the no-skill condition by \deltavsnoskill{} percentage points and the benchmark's curated skill by \deltavscurated{} percentage points, while skill evolution receives neither task rewards nor benchmark verifier outputs. In controlled ablations, using two with-skill runs instead of a matched with/without-skill pair reduces mean reward by 4.6 percentage points, while limiting revision to one round reduces it by 9.0 points relative to repeated revision. Together, these results show that behavioral differences between matched executions can provide an effective signal for revising skills when outcome supervision is unavailable.

\end{abstract}

\section{Introduction}
\label{sec:intro}

Agent skills equip large language model (LLM) agents with reusable procedures for professional tasks without updating model parameters. Under the Agent Skills specification, a skill is a package of procedural instructions that may also include scripts and reference resources~\citep{anthropic2025skills,xu2026agentskillslargelanguage}. Although curated skills improve task completion on professional agent benchmarks~\citep{li2026skillsbenchbenchmarkingagentskills}, an available skill may not meet the requirements of the current task. For example, a Git workflow skill may instruct an agent to push changes directly to the default branch, even though the repository requires feature branches and pull requests. Following the skill exactly would then violate the repository's workflow requirements. This mismatch motivates studying how to revise existing skills to meet the requirements of the current task.

Existing research on automatic skill improvement includes three relevant lines of work. The first extracts reusable procedures from past execution trajectories~\citep{wang2024voyageropenendedembodiedagent,zhao2024expelllmagentsexperiential,shinn2023reflexionlanguageagentsverbal}. The second revises existing skills and uses validation performance, ground-truth answers, or oracle judgments to decide which changes to retain~\citep{yang2026skilloptexecutivestrategyselfevolving,zhang2026coevoskillsselfevolvingagentskills,alzubi2026evoskillautomatedskilldiscovery}. The third updates skills using failures or feedback collected during deployment~\citep{zhang2026skillevolverskilllearningmetaskill,liu2026skillforgeforgingdomainspecificselfevolving,ma2026skillclawletskillsevolve}. The first line focuses on acquiring reusable procedures rather than revising an existing skill for the current task. The latter two address skill revision, but typically rely on external outcome signals to guide updates or select revisions. Obtaining these signals may require dedicated verifiers, labeled solutions, additional validation tasks, or feedback collected after deployment. These signals may be costly or unavailable for a new task, making it difficult to revise an existing skill using methods that depend on them. This motivates our central question: can an agent revise an existing skill using only the task, workspace, and observable execution traces, without external outcome supervision?

\begin{figure*}[!t]
\centering
\includegraphics[width=\textwidth]{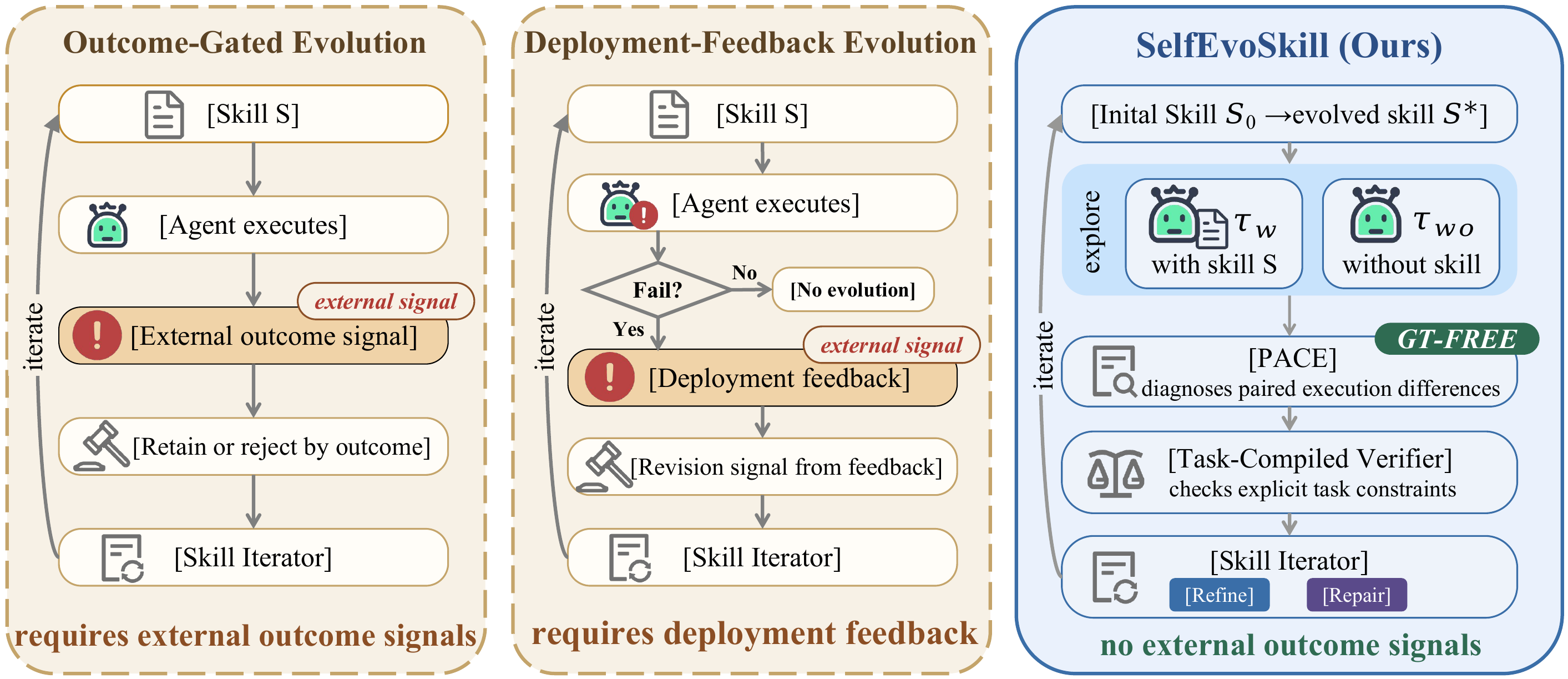}
\caption{Three paradigms for skill revision. \emph{Outcome-Gated} (left) and \emph{Deployment-Feedback} (center) rely on external outcome signals. \sysname{} (right) uses behavioral differences between matched executions with and without the skill to locate relevant skill passages and iteratively revise them, without external outcome supervision.}
\label{fig:pipeline}
\end{figure*}

Execution traces provide evidence of how an agent carries out a task, including the instructions it follows, the tools it uses, and the artifacts it produces. However, observing a problematic action in a single execution is not sufficient to determine whether the skill contributed to that action, since the agent may behave similarly without the skill. An execution without the skill therefore provides a useful reference for interpreting the behavior observed when the skill is loaded. Comparing the two executions under the same task, workspace, and agent settings helps distinguish behavior associated with skill loading from behavior that also occurs without it. In the Git example, a direct push observed in both executions provides limited evidence for attributing the action to the skill's push instruction, whereas a direct push observed only when the skill is loaded gives a more specific reason to examine that instruction. These comparisons therefore provide a basis for locating skill instructions that may contribute to problematic behavior, while subsequent executions allow the effects of revisions to be examined.

We propose \sysname{}, a framework for iterative skill revision that uses behavioral differences between matched executions as its revision signal, without task rewards, benchmark verifier outputs, or other external outcome supervision. It compares executions with and without the skill under the same task, workspace, and agent settings, then maps the observed differences to specific skill passages. This mapping connects skill instructions to their observed effects, guiding which passages to revise and which to preserve. Each revision is evaluated through subsequent executions, whose traces inform whether to retain the changes and where further revision is needed. The framework rejects changes that introduce harmful behavior, while checks derived once from the task's explicit requirements prevent revisions from violating task constraints. By connecting behavioral comparison, targeted editing, and repeated evaluation, \sysname{} enables skill adaptation using only the task, workspace, and observable execution traces. Figure~\ref{fig:pipeline} contrasts this framework with approaches that rely on external outcome supervision.

Our evaluation examines whether skill revision without external outcome supervision improves task performance and whether paired comparison and repeated revision contribute to that improvement. Across \numtasks{} containerized tasks in \numdomains{} professional domains, \sysname{} outperforms the no-skill and curated-skill baselines by \deltavsnoskill{} and \deltavscurated{} percentage points in mean task reward under the same evaluation budget (Table~\ref{tab:main}). Ablations show that comparing executions with and without the skill is more effective than using two executions that both load it, supporting the value of an execution without the skill as a reference. Further gains from repeated revision over a single round support using subsequent execution evidence to refine earlier edits. Analysis by knowledge type examines how these improvements vary with the execution evidence available for revision.

\textbf{Contributions.}
(1) \textbf{Skill revision from paired runs without outcome supervision.}
We propose \sysname{}, which uses behavioral differences
between matched executions with and without a skill to
guide revision without task rewards, benchmark verifier
outputs, or other outcome signals.
(2) \textbf{Iterative passage-level revision from execution differences.}
We link behavioral differences to passages to revise or
preserve, then use subsequent executions to evaluate edits
and guide further revision. Harmful edits are rejected,
and fixed checks enforce explicit task constraints.
(3) \textbf{Empirical gains from skill revision without outcome supervision.}
Across \numtasks{} tasks in \numdomains{} professional domains,
\sysname{} outperforms both baselines. Ablations support
paired comparison and repeated revision, while analysis
by knowledge type relates gains to available execution evidence.
\section{Related Work}
\label{sec:related}

\paragraph{Procedural memory and skill construction.}
Frozen agents can load external procedural knowledge at inference time without updating weights~\citep{yao2023reactsynergizingreasoningacting}. Voyager, ExpeL, Reflexion, Self-Refine, and Agent Workflow Memory distill reusable procedures from execution history or model feedback~\citep{wang2024voyageropenendedembodiedagent,zhao2024expelllmagentsexperiential,shinn2023reflexionlanguageagentsverbal,madaan2023selfrefineiterativerefinementselffeedback,wang2024agentworkflowmemory}. Agent Skills specify this knowledge as loadable packages~\citep{anthropic2025skills,xu2026agentskillslargelanguage}; Trace2Skill, SkillWeaver, AutoSkill, and SkillFoundry construct such packages from trajectories or other experience~\citep{ni2026trace2skilldistilltrajectorylocallessons,zheng2025skillweaverwebagentsselfimprove,yang2026autoskillexperiencedrivenlifelonglearning,shen2026skillfoundrybuildingselfevolvingagent}. SkillTTA further synthesizes a temporary skill from retrieved task-relevant trajectories at test time~\citep{wang2026skilltta}. The memory and synthesis methods above create, retrieve, or accumulate procedural knowledge; they do not isolate the incremental effect of an already available skill in a matched execution pair. We instead ask how to locally revise such a skill for the task at hand when neither task reward nor a held-out validation result is available.

\paragraph{Skill revision with outcome supervision.}
Several lines of work iteratively revise skills but retain updates using an external outcome signal (see Figure~\ref{fig:pipeline}). SkillOpt, EvoSkill, CoEvoSkills, SkillsVote, and SkillRevise use held-out validation performance, ground-truth answers, oracle judgments, success-gated evidence, or re-executed empirical utility to select updates~\citep{yang2026skilloptexecutivestrategyselfevolving,alzubi2026evoskillautomatedskilldiscovery,zhang2026coevoskillsselfevolvingagentskills,liu2026skillsvote,liu2026skillrevise}. SkillForge, SkillClaw, and SkillEvolver revise skills from deployment failures or feedback~\citep{liu2026skillforgeforgingdomainspecificselfevolving,ma2026skillclawletskillsevolve,zhang2026skillevolverskilllearningmetaskill}; RL-based approaches optimize task reward directly~\citep{wang2025reinforcementlearningselfimprovingagent,xia2026skillrlevolvingagentsrecursive,vishe2026skillr1agentskillevolution,wang2026skillsdskillconditionedselfdistillationmultiturn,shi2026skill1unifiedevolutionskillaugmented,ouyang2026skilloslearningskillcuration,zhang2026memskilllearningevolvingmemory}. These methods require external outcome signals unavailable in our setting.

\section{Method}
\label{sec:method}

\subsection{Problem Formulation}
\label{sec:method:setup}

Given a natural-language task description $T$, an initial
workspace state $W$, an initial skill $S_0$, and a frozen
agent $\pi$, our goal is to revise $S_0$ to improve task
performance without updating the agent's parameters.
For a candidate skill $S$, execution produces a trajectory
$\tau \sim \pi(\cdot \mid S,T,W)$.
Let $\mathcal{R}(\tau) \in [0,1]$ denote the task reward.
The ideal objective is
\begin{equation}
\label{eq:objective}
S^* \in \arg\max_{S}
\mathbb{E}_{\tau \sim \pi(\cdot \mid S,T,W)}
\left[\mathcal{R}(\tau)\right].
\end{equation}

We study this objective without external outcome supervision,
under the information restriction $\mathcal{C}_{\mathrm{osf}}$.
The revision procedure may access $T$, $W$, the initial and
candidate skills, and observable execution traces and artifacts.
It may not access $\mathcal{R}$ or its outputs, held-out
validation results, reference solutions, benchmark verifier
outputs, or oracle pass/fail feedback.
Checks derived solely from explicit task requirements and
observable workspace information are permitted.
Thus, $\mathcal{R}$ defines the performance objective but
provides no feedback during revision.

\subsection{\sysname{} Overview}
\label{sec:method:overview}

\sysname{} revises skills through an iterative process of
execution comparison, targeted editing, and evaluation,
without external outcome supervision.
It combines behavioral evidence from matched executions
with checks derived from explicit task requirements:
the former guides skill revision, while the latter
constrains the resulting changes.

At each iteration, the agent executes the task with and
without the current skill under the same task, initial
workspace state, and agent settings.
\textsc{PACE}, our trajectory comparison and skill diagnosis
module, examines the paired trajectories and the current
skill along four dimensions: Process Adherence, Artifact
Evidence, Consistency, and Effectiveness Delta.
It links observed behavioral differences to skill passages
that should be revised or preserved.
The Skill Iterator then applies targeted edits under
Refine or Repair, a revision mode selected by the initial
paired audit and fixed throughout the loop.
Subsequent executions assess the updated skill and
provide evidence for further revision.
The Task-Compiled Verifier, constructed before revision
begins and held fixed throughout the loop, checks compliance
with explicit task requirements.

\begin{figure*}[t]
\centering
\includegraphics[width=\textwidth]{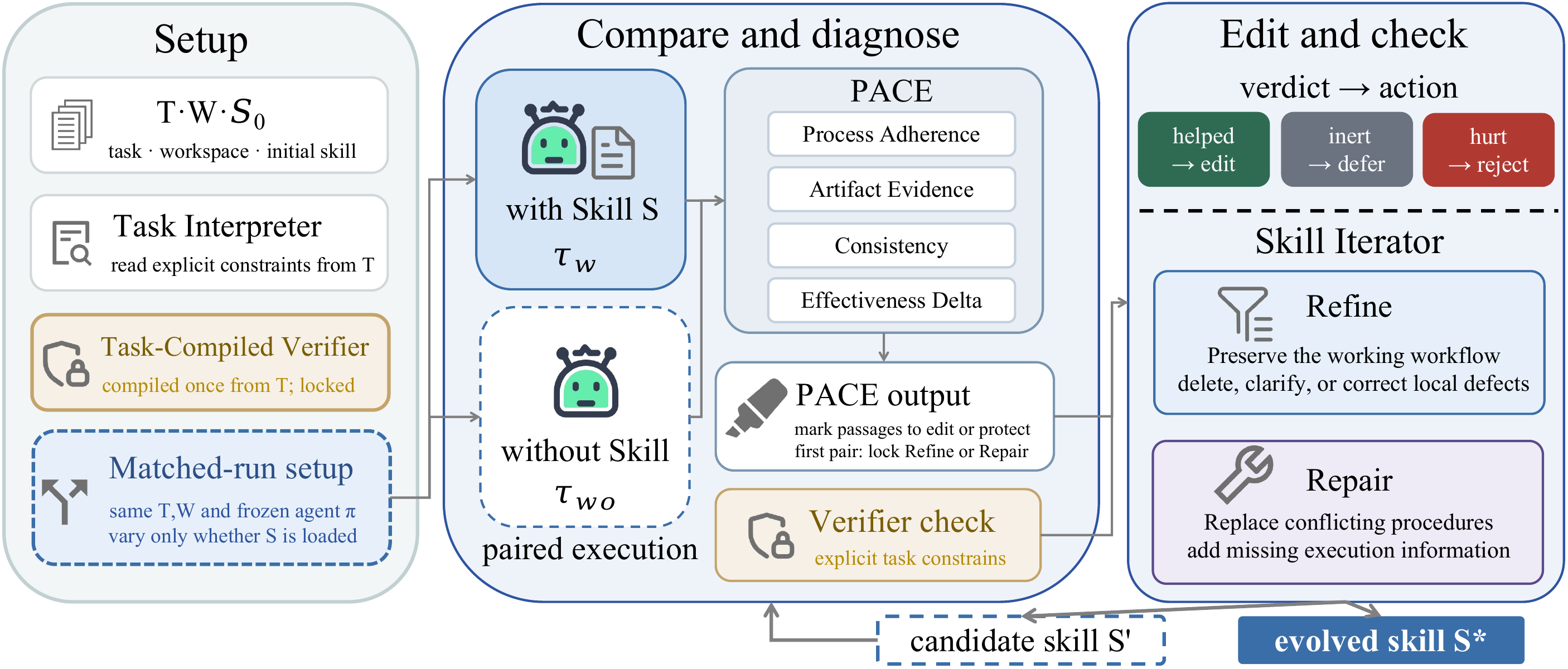}
\caption{\sysname{} overview.
\emph{Setup} (left): compile a verifier from explicit task requirements and select \emph{Refine} or \emph{Repair} through an initial paired audit. Both remain fixed.
\emph{Compare and diagnose} (center): \textsc{PACE} compares executions with and without the skill to identify passages to edit or preserve.
\emph{Edit and check} (right): the Skill Iterator edits the skill under the selected mode, and subsequent execution checks the candidate. A \emph{hurt} verdict or constraint violation triggers rejection.}
\label{fig:detail}
\end{figure*}

This process allows revision decisions to be reconsidered
as edits change the agent's behavior.
Figure~\ref{fig:detail} illustrates the workflow.
The following subsections describe \textit{paired trajectory auditing},
\textit{PACE}, \textit{refine and repair}, and \textit{the task-compiled verifier}.

\subsection{Paired Trajectory Auditing}
\label{sec:method:cpa}

\sysname{} examines the influence of the current skill through two executions of the frozen agent $\pi$: one with $S$ loaded and one without a skill. The resulting trajectories are
$\tau_w \sim \pi(\cdot \mid S,T,W)$ and
$\tau_{wo} \sim \pi(\cdot \mid \varnothing,T,W)$,
where $\varnothing$ denotes the absence of a skill. Both executions start from the same workspace state $W$
and use the same task $T$ and agent configuration.
Skill loading is the only factor deliberately varied.
The execution without the skill provides a reference
for assessing which observed behaviors may be associated
with the skill.

The audit compares observable tool calls, decisions,
and generated artifacts, without accessing the task
reward $\mathcal{R}$.
If the execution with the skill satisfies an explicit
task requirement that the execution without it does not,
the difference provides evidence of a beneficial skill
effect on that requirement.
Matching behavior provides no evidence of an incremental
effect for the behavior being compared.
Conversely, a call or artifact that conflicts with $T$
indicates a problem in execution, whose connection to
the skill must be examined using the paired trajectories
and skill instructions.
\textsc{PACE} performs this diagnosis and identifies
its implications for revision.
If the execution without the skill is too incoherent
to provide a meaningful reference, the framework
disables edits that import fragments from that trajectory.
It relies on checks of the execution with the skill
until a later pair provides an informative comparison
(Appendix~\ref{sec:appendix:impl}).

\subsection{PACE}
\label{sec:method:pace}

\textsc{PACE} translates paired execution evidence into
passage-level diagnoses.
It examines $\tau_w$, $\tau_{wo}$, and the current skill
$S$ together, since a behavioral difference alone does
not identify the instruction associated with it.
Its outputs comprise a behavioral verdict and specific
skill passages marked for revision or preservation.
These passage-level recommendations guide the
Skill Iterator.

\textsc{PACE} evaluates four complementary dimensions: \textbf{(1) Process Adherence.}
This dimension examines whether the agent follows the
steps and tool-use patterns specified in $S$, providing
evidence about which instructions are enacted during
execution. \textbf{(2) Artifact Evidence.}
This dimension checks the files produced by both runs
against explicit requirements in $T$ and the deterministic
checks provided by the Task-Compiled Verifier.
The skill may be used to investigate an observed mismatch,
but never to define the requirements against which
artifacts are assessed. \textbf{(3) Consistency.}
This dimension examines whether the workflow specified
in $S$ is compatible with the task's data schema and
output format. \textbf{(4) Effectiveness Delta.}
This dimension assesses the incremental behavioral
contribution of loading $S$ relative to the execution
without it. Observable calls, decisions, and artifacts
support a judgment of clearly positive, near-zero,
or negative contribution. This judgment concerns
observed execution behavior, not task reward or
estimated task success.

All four dimensions use only the task, permitted workspace
information, skill instructions, and observable execution
evidence. None accesses task rewards, benchmark tests,
benchmark verifier outputs, reference solutions, or
oracle pass/fail feedback.
See Appendix~\ref{sec:appendix:pace} for evaluator details.

\paragraph{Verdict aggregation and use.}
\textsc{PACE} aggregates the judgments according to
a precedence rule.
Any individual \emph{hurt} judgment takes precedence.
Otherwise, Artifact Evidence determines the verdict
when the file outputs resolve the comparison.
If they do not, a majority of the remaining judgments
determines a \emph{helped} or \emph{inert} verdict.
The verdict summarizes the behavioral evidence, while
the passage-level diagnosis identifies the corresponding
revision or preservation actions.
For the initial skill, Effectiveness Delta and the scope
of the diagnosis determine the fixed Refine and Repair modes
(\S\ref{sec:method:dual}).
For an edited candidate, a \emph{hurt} verdict triggers
rejection.

\subsection{Refine and Repair Modes}
\label{sec:method:dual}

\sysname{} distinguishes between preserving a useful
procedure and replacing an ineffective or conflicting one.
The initial paired audit selects either \emph{Refine}
or \emph{Repair}, and the selected mode remains fixed
throughout revision.

\paragraph{Mode selection.}
A clearly positive Effectiveness Delta for $S_0$ selects
Refine; a negative one selects Repair.
For near-zero judgments, local redundancy, ambiguity,
or clarifiable defects select Refine, while missing
execution information, nonlocal workflow conflicts,
or harmful procedures select Repair.
The mode remains fixed thereafter.

\paragraph{Permitted edits.}
The Skill Iterator edits passages identified by \textsc{PACE}.
Refine preserves the executed procedure while removing
unused passages, clarifying instructions, or correcting
local defects.
Repair permits removing conflicting procedures, replacing
them with fragments from $\tau_{wo}$ checked for compliance
with $T$, and then adding missing execution details.
Both modes preserve protected passages.
Repair allows all passages to be examined but respects
these protections.
The candidate is evaluated in the next execution with
the skill.

\paragraph{Initial diagnosis and candidate rejection.}
Mode selection is distinct from candidate acceptance.
The initial audit fixes the mode even when the PACE
verdict is \emph{inert}, and a negative initial diagnosis
can select Repair rather than terminate revision.
Once an edit produces a candidate, however, a subsequent
\emph{hurt} verdict rejects that candidate and restores
the most recent skill that passed the verifier.
This distinction allows an initially ineffective or
misleading skill to be repaired while preventing harmful
candidates from replacing an earlier skill.

\subsection{The Task-Compiled Verifier}
\label{sec:method:verifier}

Improvement in the paired comparison does not guarantee
compliance with explicit task requirements.
Before revision begins, \sysname{} compiles requirements
in $T$ that admit deterministic checks into a
Task-Compiled Verifier.
Workspace data $W$ is used only to resolve task-referenced
files and recompute values required by $T$, not to introduce
additional requirements.
Checks cover required files and paths, headers or schemas,
enumerated value sets, recomputable values, and companion
files.
Compilation excludes the initial skill, paired trajectories,
task rewards, benchmark tests, and benchmark verifier outputs.
The verifier remains fixed throughout revision.

The verifier checks each candidate through its subsequent
execution with the skill.
A detected constraint violation rejects the candidate
and restores the most recently accepted skill,
regardless of the PACE verdict.
If $S_0$ fails, the failure is recorded because no earlier
skill is available for rollback.
When the initial audit selects Repair, subsequent edits
may use the execution without the skill as a reference.
The verifier enforces explicit task constraints.
It neither assesses the skill's behavioral contribution
nor selects the revision mode, and it does not replace
benchmark evaluation.
See Appendices~\ref{sec:appendix:algo}
and~\ref{sec:appendix:impl} for the full algorithm
and implementation details.

\section{Experiments}
\label{sec:exp}

\subsection{Experimental Settings}
\label{sec:exp:setup}

\paragraph{Benchmark and metric.}
We evaluate on a fixed snapshot of SkillsBench containing
\numtasks{} tasks across \numdomains{} professional
domains~\citep{li2026skillsbenchbenchmarkingagentskills}.
Each task provides a curated skill, a Harbor
container~\citep{merrill2026terminalbenchbenchmarkingagentshard},
and a deterministic benchmark verifier that returns
a task reward in $[0,1]$, our evaluation metric.
See Appendix~\ref{sec:appendix:tasklevel} for task details and results.

\begin{figure}[t]
\centering
\includegraphics[width=\textwidth]{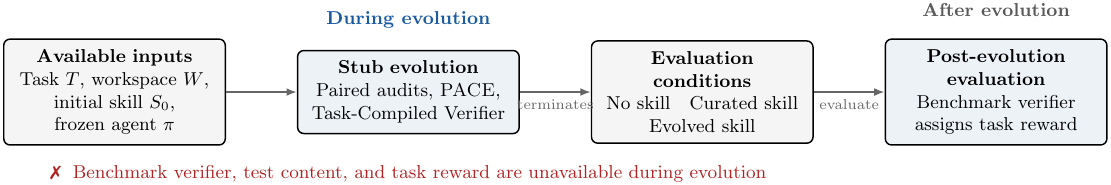}
\caption{Evaluation protocol without outcome supervision. Evolution uses task inputs and execution traces without access to benchmark verifiers, tests, or rewards. After evolution, the same agent is evaluated under no-skill, curated-skill, and evolved-skill conditions using the benchmark verifier.}
\label{fig:no-outcome-supervision-protocol}
\vspace{-10pt}
\end{figure}

\paragraph{Revision setup.}
Skill revision takes place in a stub container without
access to the benchmark's pytest verifier or test content.
The benchmark verifier is invoked only after revision
terminates.
Agent execution, PACE assessment, skill editing, and
evaluation runs all use Claude Code with Claude Opus 4.8.
Revision allows at most five paired audits per task.
All tasks use the same twelve PACE evaluator templates,
paper-mode prompts, and fixed rule for selecting the
revision mode, with no task-specific tuning.
Figure~\ref{fig:no-outcome-supervision-protocol}
illustrates the separation between revision and evaluation.

\paragraph{Baselines and evaluation protocol.}
We compare three conditions using the same frozen agent:
no skill, the benchmark's curated skill, and the skill
revised by \sysname{}.
The first two conditions each use two independent task
executions.
For \sysname{}, two independent revision runs produce
two skills, each evaluated once.
We report the higher of the two rewards for each condition,
following a common best-of-two reporting protocol.

\begin{table}[t]
\centering
\small
\setlength{\tabcolsep}{6pt}
\caption{Mean task reward (\%) on the fixed \numtasks{}-task snapshot across \numdomains{} domains, evaluated with Claude Opus 4.8. The three performance columns report mean per-task rewards in $[0,1]$, expressed as percentages. Curated skill denotes the benchmark-provided skill supplied with the task; \sysname{} denotes the evolved skill. In the \textbf{Overall} row, $N$ gives the total task count, and the performance entries are computed over all \numtasks{} tasks rather than as an unweighted mean of the eight domain rows. The last column reports the gain of \sysname{} over the curated skill. }
\label{tab:main}
\begin{tabular}{@{}l r r r r r@{}}
\toprule
Domain & $N$ & No skill & Curated skill & \sysname{} & $\Delta$ vs Curated skill (pp) \\
\midrule
Software Engineering           & 16 & 53.8 & 40.3 & \textbf{78.8} & $+38.5$ \\
Office \& White Collar         & 15 & 60.0 & 60.0 & \textbf{86.7} & $+26.7$ \\
Natural Science                & 15 & 41.6 & 73.3 & \textbf{83.7} & $+10.4$ \\
Industrial \& Physical Sys.    & 14 & 21.4 & 61.2 & \textbf{71.4} & $+10.2$ \\
Finance \& Economics           &  9 & 33.3 & \textbf{44.4} & \textbf{44.4} & $+0.0$ \\
Mathematics \& OR              &  8 & 25.0 & 35.6 & \textbf{57.5} & $+21.9$ \\
Cybersecurity                  &  6 & 41.7 & 63.8 & \textbf{66.7} & $+2.9$ \\
Media \& Content Prod.         &  6 & 33.8 & 79.2 & \textbf{83.3} & $+4.1$ \\
\midrule
\textbf{Overall} & \textbf{\numtasks{}} & \passratenoskillvalue{} & \passratecuratedvalue{} & \textbf{\passratefullvalue{}} & $\mathbf{\deltavscurated}$ \\
\bottomrule
\end{tabular}
\end{table}

\subsection{Main Results}
\label{sec:exp:main}

Table~\ref{tab:main} reports per-domain mean task reward, and Table~\ref{tab:positioning} compares \sysname{} with other skill-evolution methods on SkillsBench.

\paragraph{Paired auditing without outcome supervision improves task reward.}
Over the \numtasks{} tasks, \sysname{} attains \passratefull{}, with gains of \deltavsnoskill{} pp over the no-skill condition and \deltavscurated{} pp over the curated skill. 
The evolved skill outperforms the curated skill in seven of eight domains. 
\begin{wraptable}[12]{r}{0.46\textwidth}
\centering
\small
\setlength{\tabcolsep}{3.5pt}
\caption{Comparison with other methods on SkillsBench. External results are taken from the cited papers and were not reproduced in our harness. The values do not constitute a controlled comparison. }
\label{tab:positioning}
\begin{tabular}{@{}l l r l@{}}
\toprule
Method & Executor & Reward (\%) & \shortstack{External\\outcome\\signal} \\
\midrule
\sysname{} & Opus 4.8 & \passratefullvalue{} & None \\
CoEvoSkills & Opus 4.6 & 71.1 & Oracle \\
SkillRevise & Opus 4.7 & 51.2 & Verifier \\
\bottomrule
\end{tabular}
\end{wraptable}
In Finance \& Economics, the evolved and curated skills both attain $44.4\%$, so this condition is a tie rather than an improvement. The largest gains occur in Software Engineering ($+38.5$ pp), Office \& White Collar ($+26.7$ pp), and Mathematics \& OR ($+21.9$ pp). In Software Engineering, the curated skill is below the no-skill condition. In Office \& White Collar, it provides no gain over the no-skill condition. Mathematics \& OR has the lowest curated-skill reward among the eight domains. These conditions leave substantial room for skill revision.   (Appendix~\ref{sec:appendix:tasklevel}).

\paragraph{The evolved skill achieves comparable reward without outcome supervision.}
Table~\ref{tab:positioning} compares \sysname{} with two other skill-evolution methods evaluated on SkillsBench: CoEvoSkills~\citep{zhang2026coevoskillsselfevolvingagentskills} and SkillRevise~\citep{liu2026skillrevise}. Despite receiving no external outcome signals during evolution, \sysname{} reports a mean reward in the same performance range as these outcome-supervised methods. Table~\ref{tab:external-protocols} details the differences in task snapshots, models, harnesses, supervision signals, and reporting protocols.

\subsection{Ablation Studies}
\label{sec:exp:ablation}

We ablate four design choices in \sysname{}:
the execution without the skill as a reference,
audit-based selection between Refine and Repair,
the addition of missing execution information during Repair, and iterative revision with PACE.
All variants use GLM-5.2 as the evolution model and share the same task set and evaluation protocol, allowing controlled comparisons of these design choices. The full method achieves a mean reward of $73.0\%$ under GLM-5.2, providing evidence of effectiveness with a second evolution backbone alongside the \passratefull{} result under Claude Opus 4.8.

\paragraph{The without-skill reference provides the update signal.}
If both executions load the same skill, trajectory differences reflect sampling variation and cannot indicate whether the skill is used. We hold the procedure fixed and change only the pairing. Both conditions use two executions per iteration, matching the compute budget: one run with the skill and one without, versus two runs that both load the same skill. The comparison follows the same best-of-two reporting protocol. 
Both arms are evolved and evaluated with GLM-5.2 on all \numtasks{} tasks, rather than the Claude Opus 4.8 protocol used for Table~\ref{tab:main}. Mean reward is $64.0\%$ for the with/without pair and $59.4\%$ for two with-skill runs ($+4.6$ pp). The without-skill run provides a contrastive reference for estimating the skill's incremental effect.

\paragraph{A paired reference prevents an accidental trace from triggering an edit.}
The aggregate comparison above establishes a difference in mean reward; \texttt{grid-dispatch-operator} records the decision behind it. With the without-skill reference, the framework observes that the with-skill workflow remains optimal while the without-skill run fails load balance, so it leaves the skill unchanged. When both runs load the skill, a silent infeasibility is judged as skill harm; the resulting rewrite still passes structural checks but receives task reward $0$. This archived example illustrates how the paired reference guides the update in the GLM pairing run (Appendix~\ref{sec:appendix:case-pair}).

\begin{table}[!t]
\centering
\footnotesize
\captionsetup[subtable]{justification=centering,singlelinecheck=false}
\setlength{\tabcolsep}{2.5pt}
\caption{Ablation on \numtasks{} tasks. \textbf{(a)} Mean reward (\%) by variant; \textbf{(b)} complementary solves (\%) by fixed edit mode. All skills evolve with GLM-5.2; missing or failed runs score 0.}
\label{tab:ablation}
\begin{subtable}[t]{0.49\textwidth}
\centering
\begin{tabular}{@{}lccc@{}}
\toprule
Variant & GLM & Qwen & DS \\
\midrule
Refine-only & 62.9 & 46.1 & 38.2 \\
Repair-only & 67.0 & 50.4 & 39.7 \\
Repair, replace only & 65.1 & 44.7 & 37.4 \\
Single-round \textsc{PACE} & 64.0 & 42.7 & 35.4 \\
\midrule
Adaptive routing & \textbf{73.0} & \textbf{53.2} & \textbf{42.4} \\
\shortstack[l]{Per-task max\\(upper bound)} & 77.3 & 58.5 & 47.9 \\
\bottomrule
\end{tabular}
\caption{Mean reward by variant.}
\label{tab:ablation_variants}
\end{subtable}\hfill
\begin{subtable}[t]{0.47\textwidth}
\centering
\begin{tabular}{@{}lccc@{}}
\toprule
Category & GLM & Qwen & DS \\
\midrule
Both solve & 52.8 & 39.3 & 29.2 \\
Refine-only solves & 10.1 & 6.7 & 9.0 \\
Repair-only solves & 13.5 & 11.2 & 10.1 \\
Neither solves & 23.6 & 42.7 & 51.7 \\
\bottomrule
\end{tabular}
\caption{Complementary solves.}
\label{tab:routing-decomp}
\end{subtable}
\vspace{-15pt}
\end{table}

\paragraph{Adaptive routing preserves the complementary strengths of both modes.}
A workflow that improves execution should be preserved, whereas one with no effect or that misleads the agent may be replaced. If only one edit mode is permitted in both cases, some tasks become unrecoverable. In Table~\ref{tab:ablation} and Figure~\ref{fig:ablation_routing}, Refine-only attains $62.9\%$ on GLM and Repair-only attains $67.0\%$. Table~\ref{tab:routing-decomp} shows that Refine uniquely solves $10.1\%$ of tasks and Repair uniquely solves $13.5\%$: using only one mode forfeits the tasks that the other mode solves. Selecting the mode from the first paired audit reaches $73.0\%$ on GLM, near the per-task upper bound of $77.3\%$. When these GLM-evolved skills are evaluated on Qwen and DeepSeek without re-evolution, they reach $53.2\%$ and $42.4\%$, still above either mode alone (\S\ref{sec:method:dual}).

\paragraph{Repair benefits from adding missing execution information.}
After removing conflicting passages, Repair can add missing execution information. Allowing this operation improves the GLM reward from $65.1\%$ under replacement only to $67.0\%$ under Repair. Thus, replacement alone does not recover all execution information used by Repair.

\paragraph{Repeated audits expose additional edit opportunities.}
Restricting \textsc{PACE} to a single round attains $64.0\%$ on GLM, $9.0$ pp below the multi-round variant with the selected edit mode ($73.0\%$). A single audit round can leave actionable differences unaddressed; later rounds act after an earlier edit changes the execution trace.

\subsection{Analysis of Paired Process Evidence}
\label{sec:exp:analysis}

\begin{figure}[!t]
\centering
\begin{subfigure}[t]{0.48\textwidth}
\centering
\includegraphics[height=1.50in,keepaspectratio]{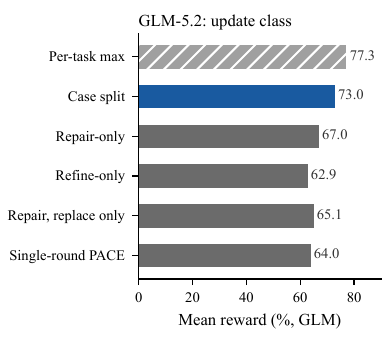}
\caption{Routing from the first paired audit.}
\label{fig:ablation_routing}
\end{subfigure}\hfill
\begin{subfigure}[t]{0.48\textwidth}
\centering
\includegraphics[height=1.50in,keepaspectratio]{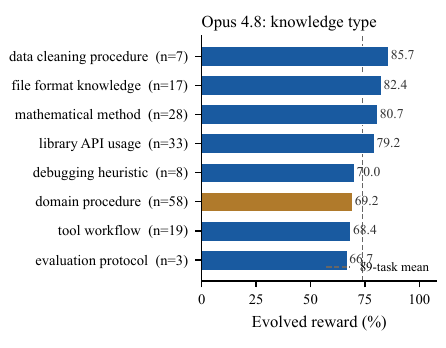}
\caption{Mean evolved reward by knowledge type.}
\label{fig:analysis_traces}
\end{subfigure}
\caption{Paired routing and performance by knowledge type.
\textbf{(a)} Audit-based mode selection outperforms either fixed mode in the GLM ablation; hatched bars show the per-task upper bound.
\textbf{(b)} Mean rewards by knowledge type on the Opus \numtasks{}-task split. Labels may overlap; gold denotes domain procedure, and the dashed line marks the overall mean.}
\label{fig:ablation_analysis}
\vspace{-10pt}
\end{figure}

\paragraph{Skills with observable execution signals benefit most from paired revision.}
Skills whose procedures specify library functions or mathematical formulas exceed the most frequent domain-procedure label (Figure~\ref{fig:analysis_traces}): library API usage attains $79.2\%$ ($n=33$) and mathematical method attains $80.7\%$ ($n=28$), compared with $69.2\%$ for domain procedure ($n=58$). For these skills, execution often exposes runtime failures or incorrect values, providing concrete signals that make relevant passages easier to diagnose. To obtain these per-type statistics, we constructed an eight-category taxonomy from the associated skills and task set. An LLM assigned one or more labels from this taxonomy to \numtasks{} tasks, and two authors reviewed and corrected every assignment. Labels overlap, and $n$ denotes the number of tasks bearing a label. Per-type statistics appear in Appendix~\ref{sec:appendix:tasklevel}.

\paragraph{A case study identifies a conflicting instruction.}
In \texttt{paper-anonymizer}, the loaded skill aborts a redaction unless more than $80\%$ of the source text remains, and its verification code enforces a $0.7$ floor, although the task specification requires only that at least $60\%$ ($0.6$) be retained. The over-strict guard therefore rejects valid redactions. \textsc{PACE} marks that guard as an edit target (ST-002/ST-003) and unwraps a generic function wrapper (ST-001), protecting the remaining sections; the repaired skill sets both thresholds to $0.6$, and the task's evolved-skill reward rises from $0$ to $1$ (Repair). This case records a passage-level diagnosis for a conflicting instruction (Appendix~\ref{sec:appendix:case-pace}).

\paragraph{Two case studies trace protection and repair decisions.}
The following recorded cases trace two distinct decisions. In \texttt{software-dependency-audit}, Refine removes only unused or conflicting passages from a workflow supported by the paired runs: it removes tutorial prose and hardcoded paths, cutting about $35\%$ of lines, and the evolved reward is $1.0$. In \texttt{data-to-d3}, the first with-skill run passes 12 of 13 checks but violates the task's exact column-label requirement, whereas the without-skill run passes all 13. The locked verifier prevents that failing state from being accepted, while the Repair mode selected by the same audit remains available. By the final archived iteration, the evolved skill passes all 13 checks after shrinking from 188 lines to 50, and its reward rises from $0$ to $1.0$. This case illustrates the verifier's role within a repair trajectory. Appendix~\ref{sec:appendix:cases} gives complete records, and Appendix~\ref{sec:appendix:length} reports the resulting package-level changes.

\paragraph{Failure analysis identifies execution as the largest residual bottleneck.}
We analyzed 25 non-passing cases using their final evaluation outputs, evolved skills, and audit records. In 14 cases ($56.0\%$), the evolved skill contained the relevant task guidance, but the frozen agent failed to follow it correctly in the final output. Five cases ($20.0\%$) resulted from incorrect or harmful skill revisions, four ($16.0\%$) from PACE failing to localize an effective edit, and two ($8.0\%$) remained uncertain because the available records did not support a unique attribution. Execution was therefore the largest observed failure source, while nine cases directly point to opportunities for improving passage localization and skill editing.

\paragraph{Limitations of paired process evidence.}
Execution errors by the frozen agent account for the largest share of non-passing cases ($56.0\%$): paired process evidence can identify passage-level defects in the skill but cannot compensate when the agent fails to follow a correctly revised procedure. The remaining nine attributable cases---incorrect revisions ($20.0\%$) and localization failures ($16.0\%$)---point to opportunities for improving how trajectory differences are connected to skill passages and how edits are applied. Appendix~\ref{sec:appendix:tasklevel} provides the task-level breakdowns.

\section{Conclusion}
\label{sec:conclusion}

We present \sysname{}, which uses behavioral differences between matched executions with and without a skill to guide iterative passage-level revision without external outcome supervision. A fixed verifier enforces explicit
task requirements. Across \numtasks{} tasks in \numdomains{} professional domains, \sysname{} achieves a mean reward of \passratefull{}, outperforming both baselines. Ablations support paired comparison and repeated revision,
showing the value of execution evidence for skill adaptation without external outcome supervision. This approach offers a practical way to adapt existing skills when task rewards and reference solutions are unavailable.

\section*{AI Use Statement}

We used generative AI tools for two purposes in preparing this submission: polishing the English wording and grammar of the manuscript and assisting with scripts for experiments and data processing. In these supporting roles, generative AI tools were not used to generate research ideas, design the method or evaluation protocol, or produce, select, or interpret any reported numerical results. All AI-assisted text was reviewed and edited by the authors and checked for consistency with the underlying results. All AI-assisted code was reviewed and tested by at least two authors, who also inspected the recorded execution logs to check its behavior. Generative models are also used within the framework under study, serving as the frozen agent and the \textsc{PACE} evaluators. This use is part of the method and is described in \S\ref{sec:method} and Appendix~\ref{sec:appendix:pace}. We take responsibility for the final content of this work, including all text, claims, and artifacts produced with the aid of generative AI.

\section*{Ethics Statement}

This work involves no human subjects and collects no personally identifiable information. Every execution runs inside a containerized sandbox on the benchmark's own workspace data, and the models and the benchmark we use are employed under their respective licenses and intended-use terms. One risk is specific to our setting: because \sysname{} rewrites procedural instructions that a frozen agent then executes, an evolved skill could in principle encode an unsafe procedure. Two mechanisms address this. \textsc{PACE} includes a safety-compliance evaluator that vetoes an update on a critical violation, such as unsafe paths, exposed credentials, or destruction of content the task requires to be preserved (Appendix~\ref{sec:appendix:pace}); and the Task-Compiled Verifier rejects updates that violate explicit task-grounded constraints. Neither mechanism substitutes for human review before an evolved skill is deployed outside a sandbox. The authors declare no conflicts of interest.

\section*{Reproducibility Statement}

The problem setting and the constraint on revision without outcome supervision are stated formally in \S\ref{sec:method:setup}. The evolution loop, including the convergence criteria and locked initial routing rule, is given as pseudocode in Appendix~\ref{sec:appendix:algo}; the complete fixed \textsc{PACE} evaluator inventory is in Appendix~\ref{sec:appendix:pace}; and implementation details, including model versions, the iteration and edit budgets, and the task-compiled verifier boundary, are in Appendix~\ref{sec:appendix:impl}. The frozen task manifest and per-task rewards in Appendix~\ref{sec:appendix:tasklevel} identify the evaluation snapshot and support recomputation of Table~\ref{tab:main}. Six worked cases in Appendix~\ref{sec:appendix:cases} give the per-iteration decisions and the resulting skill edits. We will release the implementation, paper profile, paper-mode prompts, static evaluator templates, frozen manifest, selection and budget rules, launch scripts, audit-record schema, and evolved skill packages.

\bibliography{references}

@misc{zhang2026coevoskillsselfevolvingagentskills,
      title={CoEvoSkills: Self-Evolving Agent Skills via Co-Evolutionary Verification},
      author={Hanrong Zhang and Shicheng Fan and Henry Peng Zou and Yankai Chen and Zhenting Wang and Jiayu Zhou and Chengze Li and Wei-Chieh Huang and Yifei Yao and Kening Zheng and Xue Liu and Xiaoxiao Li and Philip S. Yu},
      year={2026},
      eprint={2604.01687},
      archivePrefix={arXiv},
      primaryClass={cs.AI},
      url={https://arxiv.org/abs/2604.01687},
}

@misc{zhang2026skillevolverskilllearningmetaskill,
      title={SkillEvolver: Skill Learning as a Meta-Skill},
      author={Genrui Zhang and Erle Zhu and Jinfeng Zhou and Caiyan Jia and Hongning Wang},
      year={2026},
      eprint={2605.10500},
      archivePrefix={arXiv},
      primaryClass={cs.AI},
      url={https://arxiv.org/abs/2605.10500},
}

@misc{ni2026trace2skilldistilltrajectorylocallessons,
      title={Trace2Skill: Distill Trajectory-Local Lessons into Transferable Agent Skills},
      author={Jingwei Ni and Yihao Liu and Xinpeng Liu and Yutao Sun and Mengyu Zhou and Pengyu Cheng and Dexin Wang and Xiaoxi Jiang and Guanjun Jiang},
      year={2026},
      eprint={2603.25158},
      archivePrefix={arXiv},
      primaryClass={cs.AI},
      url={https://arxiv.org/abs/2603.25158},
}

@misc{ma2026skillclawletskillsevolve,
      title={SkillClaw: Let Skills Evolve Collectively with Agentic Evolver},
      author={Ziyu Ma and Shidong Yang and Yuxiang Ji and Xucong Wang and Yong Wang and Yiming Hu and Tongwen Huang and Xiangxiang Chu},
      year={2026},
      eprint={2604.08377},
      archivePrefix={arXiv},
      primaryClass={cs.AI},
      url={https://arxiv.org/abs/2604.08377},
}

@misc{yang2026skilloptexecutivestrategyselfevolving,
      title={SkillOpt: Executive Strategy for Self-Evolving Agent Skills},
      author={Yifan Yang and Ziyang Gong and Weiquan Huang and Qihao Yang and Ziwei Zhou and Zisu Huang and Yan Li and Xuemei Gao and Qi Dai and Bei Liu and Kai Qiu and Yuqing Yang and Dongdong Chen and Xue Yang and Chong Luo},
      year={2026},
      eprint={2605.23904},
      archivePrefix={arXiv},
      primaryClass={cs.AI},
      url={https://arxiv.org/abs/2605.23904},
}

@misc{alzubi2026evoskillautomatedskilldiscovery,
      title={EvoSkill: Automated Skill Discovery for Multi-Agent Systems},
      author={Salaheddin Alzubi and Noah Provenzano and Jaydon Bingham and Weiyuan Chen and Tu Vu},
      year={2026},
      eprint={2603.02766},
      archivePrefix={arXiv},
      primaryClass={cs.AI},
      url={https://arxiv.org/abs/2603.02766},
}

@misc{liu2026skillforgeforgingdomainspecificselfevolving,
      title={SkillForge: Forging Domain-Specific, Self-Evolving Agent Skills in Cloud Technical Support},
      author={Xingyan Liu and Xiyue Luo and Linyu Li and Ganghong Huang and Jianfeng Liu and Honglin Qiao},
      year={2026},
      eprint={2604.08618},
      archivePrefix={arXiv},
      primaryClass={cs.IR},
      doi={10.1145/3805712.3808466},
      url={https://arxiv.org/abs/2604.08618},
}

@misc{shen2026skillfoundrybuildingselfevolvingagent,
      title={SkillFoundry: Building Self-Evolving Agent Skill Libraries from Heterogeneous Scientific Resources},
      author={Shuaike Shen and Wenduo Cheng and Mingqian Ma and Alistair Turcan and Martin Jinye Zhang and Jian Ma},
      year={2026},
      eprint={2604.03964},
      archivePrefix={arXiv},
      primaryClass={cs.AI},
      url={https://arxiv.org/abs/2604.03964},
}

@misc{yang2026autoskillexperiencedrivenlifelonglearning,
      title={AutoSkill: Experience-Driven Lifelong Learning via Skill Self-Evolution},
      author={Yutao Yang and Junsong Li and Qianjun Pan and Bihao Zhan and Yuxuan Cai and Lin Du and Jie Zhou and Kai Chen and Qin Chen and Xin Li and Bo Zhang and Liang He},
      year={2026},
      eprint={2603.01145},
      archivePrefix={arXiv},
      primaryClass={cs.AI},
      url={https://arxiv.org/abs/2603.01145},
}

@misc{wang2026skilltta,
      title={Skills on the Fly: Test-Time Adaptive Skill Synthesis for {LLM} Agents},
      author={Jingxing Wang and Chenyu Zhou and Zhihui Fu and Jun Wang and Weiwen Liu and Weinan Zhang and Jianghao Lin},
      year={2026},
      eprint={2605.16986},
      archivePrefix={arXiv},
      primaryClass={cs.CL},
      url={https://arxiv.org/abs/2605.16986},
}

@misc{liu2026skillsvote,
      title={SkillsVote: Lifecycle Governance of Agent Skills from Collection, Recommendation to Evolution},
      author={Hongyi Liu and Haoyan Yang and Tao Jiang and Bo Tang and Feiyu Xiong and Yuyu Luo and Zhiyu Li},
      year={2026},
      eprint={2605.18401},
      archivePrefix={arXiv},
      primaryClass={cs.CL},
      url={https://arxiv.org/abs/2605.18401},
}

@misc{liu2026skillrevise,
      title={SkillRevise: Improving {LLM}-Authored Agent Skills via Trace-Conditioned Skill Revision},
      author={Yuxuan Liu and Zhaochen Su and Lingyun Xie and Yuhao Zhang and Qing Zong and Jiahe Guo and Zhongwei Xie and Yiyan Ji and Yauwai Yim and Hongyu Luo and Xiyu Ren and Ruan Chenyu and Haoran Li and Yangqiu Song},
      year={2026},
      eprint={2606.01139},
      archivePrefix={arXiv},
      primaryClass={cs.AI},
      url={https://arxiv.org/abs/2606.01139},
}

@misc{xia2026skillrlevolvingagentsrecursive,
      title={SkillRL: Evolving Agents via Recursive Skill-Augmented Reinforcement Learning},
      author={Peng Xia and Jianwen Chen and Hanyang Wang and Jiaqi Liu and Kaide Zeng and Yu Wang and Siwei Han and Yiyang Zhou and Xujiang Zhao and Haifeng Chen and Zeyu Zheng and Cihang Xie and Huaxiu Yao},
      year={2026},
      eprint={2602.08234},
      archivePrefix={arXiv},
      primaryClass={cs.LG},
      url={https://arxiv.org/abs/2602.08234},
}

@misc{vishe2026skillr1agentskillevolution,
      title={Skill-R1: Agent Skill Evolution via Reinforcement Learning},
      author={Yash Vishe and Rohan Surana and Xunyi Jiang and Zihan Huang and Xintong Li and Nikki Lijing Kuang and Tong Yu and Ryan A. Rossi and Jingbo Shang and Julian McAuley and Junda Wu},
      year={2026},
      eprint={2605.09359},
      archivePrefix={arXiv},
      primaryClass={cs.LG},
      url={https://arxiv.org/abs/2605.09359},
}

@misc{wang2026skillsdskillconditionedselfdistillationmultiturn,
      title={Skill-SD: Skill-Conditioned Self-Distillation for Multi-turn LLM Agents},
      author={Hao Wang and Guozhi Wang and Han Xiao and Yufeng Zhou and Yue Pan and Jichao Wang and Ke Xu and Yafei Wen and Xiaohu Ruan and Xiaoxin Chen and Honggang Qi},
      year={2026},
      eprint={2604.10674},
      archivePrefix={arXiv},
      primaryClass={cs.LG},
      url={https://arxiv.org/abs/2604.10674},
}

@misc{wang2025reinforcementlearningselfimprovingagent,
      title={Reinforcement Learning for Self-Improving Agent with Skill Library},
      author={Jiongxiao Wang and Qiaojing Yan and Yawei Wang and Yijun Tian and Soumya Smruti Mishra and Zhichao Xu and Megha Gandhi and Panpan Xu and Lin Lee Cheong},
      year={2025},
      eprint={2512.17102},
      archivePrefix={arXiv},
      primaryClass={cs.AI},
      url={https://arxiv.org/abs/2512.17102},
}

@misc{shi2026skill1unifiedevolutionskillaugmented,
      title={Skill1: Unified Evolution of Skill-Augmented Agents via Reinforcement Learning},
      author={Yaorui Shi and Yuxin Chen and Zhengxi Lu and Yuchun Miao and Shugui Liu and Qi Gu and Xunliang Cai and Xiang Wang and An Zhang},
      year={2026},
      eprint={2605.06130},
      archivePrefix={arXiv},
      primaryClass={cs.AI},
      url={https://arxiv.org/abs/2605.06130},
}

@misc{ouyang2026skilloslearningskillcuration,
      title={SkillOS: Learning Skill Curation for Self-Evolving Agents},
      author={Siru Ouyang and Jun Yan and Yanfei Chen and Rujun Han and Zifeng Wang and Bhavana Dalvi Mishra and Rui Meng and Chun-Liang Li and Yizhu Jiao and Kaiwen Zha and Maohao Shen and Vishy Tirumalashetty and George Lee and Jiawei Han and Tomas Pfister and Chen-Yu Lee},
      year={2026},
      eprint={2605.06614},
      archivePrefix={arXiv},
      primaryClass={cs.AI},
      url={https://arxiv.org/abs/2605.06614},
}

@inproceedings{agrawal2026gepareflectivepromptevolution,
      title={GEPA: Reflective Prompt Evolution Can Outperform Reinforcement Learning},
      author={Lakshya A Agrawal and Shangyin Tan and Dilara Soylu and Noah Ziems and Rishi Khare and Krista Opsahl-Ong and Arnav Singhvi and Herumb Shandilya and Michael J Ryan and Meng Jiang and Christopher Potts and Koushik Sen and Alexandros G. Dimakis and Ion Stoica and Dan Klein and Matei Zaharia and Omar Khattab},
      booktitle={International Conference on Learning Representations (ICLR)},
      year={2026},
      note={Oral},
      eprint={2507.19457},
      archivePrefix={arXiv},
      primaryClass={cs.CL},
      url={https://arxiv.org/abs/2507.19457},
}

@inproceedings{zhang2026agenticcontextengineeringevolving,
      title={Agentic Context Engineering: Evolving Contexts for Self-Improving Language Models},
      author={Qizheng Zhang and Changran Hu and Shubhangi Upasani and Boyuan Ma and Fenglu Hong and Vamsidhar Kamanuru and Jay Rainton and Chen Wu and Mengmeng Ji and Hanchen Li and Urmish Thakker and James Zou and Kunle Olukotun},
      booktitle={International Conference on Learning Representations (ICLR)},
      year={2026},
      eprint={2510.04618},
      archivePrefix={arXiv},
      primaryClass={cs.LG},
      url={https://arxiv.org/abs/2510.04618},
}

@misc{yuksekgonul2024textgradautomaticdifferentiationtext,
      title={TextGrad: Automatic ``Differentiation'' via Text},
      author={Mert Yuksekgonul and Federico Bianchi and Joseph Boen and Sheng Liu and Zhi Huang and Carlos Guestrin and James Zou},
      year={2024},
      eprint={2406.07496},
      archivePrefix={arXiv},
      primaryClass={cs.CL},
      url={https://arxiv.org/abs/2406.07496},
}

@misc{khattab2023dspycompilingdeclarativelanguage,
      title={DSPy: Compiling Declarative Language Model Calls into Self-Improving Pipelines},
      author={Omar Khattab and Arnav Singhvi and Paridhi Maheshwari and Zhiyuan Zhang and Keshav Santhanam and Sri Vardhamanan and Saiful Haq and Ashutosh Sharma and Thomas T. Joshi and Hanna Moazam and Heather Miller and Matei Zaharia and Christopher Potts},
      year={2023},
      eprint={2310.03714},
      archivePrefix={arXiv},
      primaryClass={cs.CL},
      url={https://arxiv.org/abs/2310.03714},
}

@misc{choudhury2025processrewardmodelsllm,
      title={Process Reward Models for LLM Agents: Practical Framework and Directions},
      author={Sanjiban Choudhury},
      year={2025},
      eprint={2502.10325},
      archivePrefix={arXiv},
      primaryClass={cs.LG},
      url={https://arxiv.org/abs/2502.10325},
}

@misc{li2026skillsbenchbenchmarkingagentskills,
      title={SkillsBench: Benchmarking How Well Agent Skills Work Across Diverse Tasks},
      author={Xiangyi Li and Wenbo Chen and Yimin Liu and Shenghan Zheng and Xiaokun Chen and Yifeng He and Yubo Li and Bingran You and Haotian Shen and Jiankai Sun and Shuyi Wang and Binxu Li and Qunhong Zeng and Di Wang and Xuandong Zhao and Yuanli Wang and Roey Ben Chaim and Zonglin Di and Yipeng Gao and Junwei He and Yizhuo He and Liqiang Jing and Luyang Kong and Xin Lan and Jiachen Li and Songlin Li and Yijiang Li and Yueqian Lin and Xinyi Liu and Xuanqing Liu and Haoran Lyu and Ze Ma and Bowei Wang and Runhui Wang and Tianyu Wang and Wengao Ye and Yue Zhang and Hanwen Xing and Yiqi Xue and Steven Dillmann and Han-chung Lee},
      year={2026},
      eprint={2602.12670},
      archivePrefix={arXiv},
      primaryClass={cs.AI},
      url={https://arxiv.org/abs/2602.12670},
}

@misc{merrill2026terminalbenchbenchmarkingagentshard,
      title={Terminal-Bench: Benchmarking Agents on Hard, Realistic Tasks in Command Line Interfaces},
      author={Mike A. Merrill and Alexander G. Shaw and Nicholas Carlini and Boxuan Li and Harsh Raj and others},
      year={2026},
      eprint={2601.11868},
      archivePrefix={arXiv},
      primaryClass={cs.SE},
      url={https://arxiv.org/abs/2601.11868},
}

@article{wang2024voyageropenendedembodiedagent,
      title={Voyager: An Open-Ended Embodied Agent with Large Language Models},
      author={Guanzhi Wang and Yuqi Xie and Yunfan Jiang and Ajay Mandlekar and Chaowei Xiao and Yuke Zhu and Linxi Fan and Anima Anandkumar},
      journal={Transactions on Machine Learning Research},
      year={2024},
}

@inproceedings{shinn2023reflexionlanguageagentsverbal,
      title={Reflexion: Language Agents with Verbal Reinforcement Learning},
      author={Noah Shinn and Federico Cassano and Edward Berman and Ashwin Gopinath and Karthik Narasimhan and Shunyu Yao},
      booktitle={Advances in Neural Information Processing Systems (NeurIPS)},
      volume={36},
      year={2023},
}

@inproceedings{yao2023reactsynergizingreasoningacting,
      title={ReAct: Synergizing Reasoning and Acting in Language Models},
      author={Shunyu Yao and Jeffrey Zhao and Dian Yu and Nan Du and Izhak Shafran and Karthik Narasimhan and Yuan Cao},
      booktitle={International Conference on Learning Representations (ICLR)},
      year={2023},
}

@misc{wang2024agentworkflowmemory,
      title={Agent Workflow Memory},
      author={Zora Zhiruo Wang and Jiayuan Mao and Daniel Fried and Graham Neubig},
      year={2024},
      eprint={2409.07429},
      archivePrefix={arXiv},
      primaryClass={cs.CL},
      url={https://arxiv.org/abs/2409.07429},
}

@inproceedings{lightman2024letsverifystepbystep,
      title={Let's Verify Step by Step},
      author={Hunter Lightman and Vineet Kosaraju and Yuri Burda and Harrison Edwards and Bowen Baker and Teddy Lee and Jan Leike and John Schulman and Ilya Sutskever and Karl Cobbe},
      booktitle={International Conference on Learning Representations (ICLR)},
      year={2024},
      eprint={2305.20050},
      archivePrefix={arXiv},
      primaryClass={cs.LG},
      url={https://arxiv.org/abs/2305.20050},
}

@inproceedings{zhao2024expelllmagentsexperiential,
      title={ExpeL: LLM Agents Are Experiential Learners},
      author={Andrew Zhao and Daniel Huang and Quentin Xu and Matthieu Lin and Yong-Jin Liu and Gao Huang},
      booktitle={AAAI Conference on Artificial Intelligence},
      year={2024},
}

@inproceedings{madaan2023selfrefineiterativerefinementselffeedback,
      title={Self-Refine: Iterative Refinement with Self-Feedback},
      author={Aman Madaan and Niket Tandon and Prakhar Gupta and Skyler Hallinan and Luyu Gao and Sarah Wiegreffe and Uri Alon and Nouha Dziri and Shrimai Prabhumoye and Yiming Yang and Shashank Gupta and Bodhisattwa Prasad Majumder and Katherine Hermann and Sean Welleck and Amir Yazdanbakhsh and Peter Clark},
      booktitle={Advances in Neural Information Processing Systems (NeurIPS)},
      volume={36},
      year={2023},
}

@misc{anthropic2025skills,
      title={Agent Skills},
      author={{Anthropic}},
      year={2025},
      howpublished={\url{https://code.claude.com/docs/en/skills}},
}

@misc{zheng2025skillweaverwebagentsselfimprove,
      title={SkillWeaver: Web Agents can Self-Improve by Discovering and Honing Skills},
      author={Boyuan Zheng and Michael Y. Fatemi and Xiaolong Jin and Zora Zhiruo Wang and Apurva Gandhi and Yueqi Song and Yu Gu and Jayanth Srinivasa and Gaowen Liu and Graham Neubig and Yu Su},
      year={2025},
      eprint={2504.07079},
      archivePrefix={arXiv},
      primaryClass={cs.AI},
      url={https://arxiv.org/abs/2504.07079},
}

@misc{xu2026agentskillslargelanguage,
      title={Agent Skills for Large Language Models: Architecture, Acquisition, Security, and the Path Forward},
      author={Renjun Xu and Yang Yan},
      year={2026},
      eprint={2602.12430},
      archivePrefix={arXiv},
      primaryClass={cs.MA},
      url={https://arxiv.org/abs/2602.12430},
}

@misc{zhang2026memskilllearningevolvingmemory,
      title={MemSkill: Learning and Evolving Memory Skills for Self-Evolving Agents},
      author={Haozhen Zhang and Quanyu Long and Jianzhu Bao and Tao Feng and Weizhi Zhang and Haodong Yue and Wenya Wang},
      year={2026},
      eprint={2602.02474},
      archivePrefix={arXiv},
      primaryClass={cs.CL},
      url={https://arxiv.org/abs/2602.02474},
}
\bibliographystyle{iclr2027_conference}

\appendix
\raggedbottom
\clearpage
\begin{center}
{\Large\bfseries Appendix}
\end{center}
\vspace{0.8em}

\noindent
{\small
\setlength{\parindent}{0pt}\setlength{\parskip}{3pt}
A.~Implementation Details \dotfill~\pageref{sec:appendix:impl}\par
B.~Related Work 
\dotfill~\pageref{sec:appendix:related}\par
C.~Evolution Loop Algorithm \dotfill~\pageref{sec:appendix:algo}\par
D.~\textsc{PACE} Evaluator Inventory \dotfill~\pageref{sec:appendix:pace}\par
E.~Supplementary Evidence and Case Studies \dotfill~\pageref{sec:appendix:cases}\par
\hspace*{1.6em}\footnotesize
E.1~Case A: \texttt{software-dependency-audit} (Refine, primary) \dotfill~\pageref{sec:appendix:case-a}\par
\hspace*{1.6em}E.2~Case B: \texttt{data-to-d3} (Repair, primary) \dotfill~\pageref{sec:appendix:case-b}\par
\hspace*{1.6em}E.3~Case C: \texttt{lab-unit-harmonization} (Refine) \dotfill~\pageref{sec:appendix:case-c}\par
\hspace*{1.6em}E.4~Case D: \texttt{exceltable-in-ppt} (Repair) \dotfill~\pageref{sec:appendix:case-d}\par
\hspace*{1.6em}E.5~Case E: \texttt{paper-anonymizer} (Repair; passage-level diagnosis) \dotfill~\pageref{sec:appendix:case-pace}\par
\hspace*{1.6em}E.6~Case F: \texttt{grid-dispatch-operator} (Refine; pairing contrast) \dotfill~\pageref{sec:appendix:case-pair}\par
\hspace*{1.6em}E.7~Structural Analysis Details \dotfill~\pageref{sec:appendix:length}\par
\normalsize
F.~Task Manifest, Per-Task Results, and Label Statistics \dotfill~\pageref{sec:appendix:tasklevel}\par
}
\clearpage

\section{Implementation Details}
\label{sec:appendix:impl}

\paragraph{Domain consolidation.}
The frozen snapshot's 11 fine-grained domains are consolidated into \numdomains{} by merging small domains and renaming one: \emph{Healthcare} into \emph{Natural Science}; \emph{Energy}, \emph{Manufacturing}, and \emph{Robotics} into \emph{Industrial \& Physical Systems}; and \emph{Mathematics} renamed to \emph{Mathematics \& OR}.

\paragraph{Reporting two independent runs.}
For each task, we conduct two independent evolution runs, yielding two evolved skills that are each evaluated once by the benchmark verifier after evolution terminates. We report the higher of these two task rewards, following the fixed best-of-two protocol in \S\ref{sec:exp:setup}. A full evolution run is lengthy and can be affected by stochastic agent execution and transient API or connection failures. Using two prespecified runs reduces sensitivity to these factors. Missing, timed-out, or aborted runs receive a reward of zero. No additional runs are launched after observing benchmark rewards.

\paragraph{Protocol differences in the cross-paper comparison.}
Because rerunning CoEvoSkills and SkillRevise under a common harness exceeds our available compute budget, Table~\ref{tab:positioning} uses the results reported in their respective papers. Table~\ref{tab:external-protocols} summarizes the differences relevant to interpreting those values. The task snapshots, models, harnesses, supervision signals, and across-run aggregation differ, so the comparison provides contextual positioning rather than a controlled ranking.

\begin{table}[H]
\centering
\footnotesize
\caption{Protocol differences behind the contextual comparison in Table~\ref{tab:positioning}. CoEvoSkills describes an 87-task evaluation set and reports evolution statistics for 86 tasks. SkillRevise reports 44 successes among 86 SkillsBench tasks under the Opus 4.7 executor. Across-run reporting is distinct from candidate selection within an evolution or revision run.}
\label{tab:external-protocols}
\setlength{\tabcolsep}{3pt}
\begin{tabular}{@{}p{0.18\textwidth} p{0.25\textwidth} p{0.25\textwidth} p{0.25\textwidth}@{}}
\toprule
Dimension & \sysname{} & CoEvoSkills & SkillRevise \\
\midrule
Task snapshot & 89 tasks & 87 evaluated; evolution statistics for 86 & 86 tasks \\
Evolution / revision model & Opus 4.8 & Opus 4.6 & GPT-5.5 \\
Executor / harness & Opus 4.8 / Claude Code & Opus 4.6 / Claude Code & Opus 4.7 / Claude Code \\
Evolution signal & No external outcome signal & Surrogate-verifier feedback and oracle pass/fail & Verifier-derived utility \\
Across-run reporting & Per-task best of two independent evolution runs & Mean over five independent runs & One reported aggregate (44/86); no repeated-run aggregation reported \\
Within-run candidate selection & PACE-guided editing with verifier rollback & Iterative co-evolutionary verification & Best-observed selection over up to three revision rounds \\
\bottomrule
\end{tabular}
\end{table}

\paragraph{Verdict aggregation.}
A single \emph{skill\_hurt} report from any evaluator vetoes the update (\S\ref{sec:method:pace}). Absent any hurt signal, the Artifact Evidence evaluator (\emph{eval-output-evidence-check}) is decisive when task-grounded file checks resolve the comparison; skill text may support attribution but never supplies the checked requirement. Any remaining disagreement is settled by majority vote into \emph{skill\_helped} or \emph{skill\_inert}.

\paragraph{First paired audit.}
The first paired audit of $S_0$ determines which edit mode is permitted (\S\ref{sec:method:dual}) using the Effectiveness Delta dimension of PACE. The same traces constitute the first iteration's diagnosis. Subsequent iterations do not recompute the edit mode. A \emph{skill\_hurt} verdict vetoes the current update and does not change the locked mode. Effectiveness Delta returns a categorical judgment: clearly positive, near zero, or negative. A clearly positive judgment selects Refine and a negative judgment selects Repair. For a near-zero judgment, local redundancy, ambiguity, or a clarifiable defect selects Refine, whereas missing execution information, a nonlocal workflow conflict, or a harmful procedure selects Repair. This edit-mode choice is derived from paired process evidence, not from $\mathcal{R}$ or from the task reward of the initial skill.

\paragraph{Task-Compiled Verifier generation.}
Compilation precedes all paired executions and proceeds in two stages. First, an LLM-based task interpreter maps each explicit requirement in $T$ to a structured specification and classifies it as an output-structure check, an invariant, a value recomputable from $W$, a required companion artifact, or not programmatically verifiable. Workspace data supplies only paths and values referenced by $T$ and contributes no new requirement. Second, an LLM-based compiler maps every eligible requirement to a named deterministic check. Expected values are read or recomputed from $W$ at verification time; hard-coded answers are prohibited. A requirement that cannot be traced to $T$ and checked without a reference solution is excluded rather than approximated.

Before paired evolution begins, the generated verifier must fail on an empty output directory, emit a structured result for every admitted requirement, and record the constraint map and requirement identifiers. A verifier that passes these validation steps is locked and reused unchanged throughout evolution; a run without a valid locked verifier does not proceed. No task-specific verifier is manually modified. Initial-skill text, paired traces, evaluator reports, task rewards, benchmark tests, reference solutions, and benchmark verifier outputs are excluded from compilation. This conservative admission rule makes a failed check evidence of a violated explicit requirement rather than a judgment of overall task quality.

\paragraph{Skill Iterator edit operations.}
Given the \texttt{surgery\_targets} and \texttt{protected\_segments} from PACE together with the paired trajectory evidence, the Skill Iterator applies edits in a fixed priority order: (i) remove or correct any passage flagged as harmful; (ii) for Refine, delete or clarify redundant or conflicting passages and correct local defects, without replacing the prescribed workflow; (iii) for Repair, replace a harmful passage with a verified alternative extracted from $\tau_{wo}$, then supply missing execution information. Every edit must target a specific surgery target and be grounded in the paired-trajectory difference, and protected segments are never altered. Relative to a baseline of $B$ lines, Refine may add at most $0.1B$ net lines per round and Repair at most $0.2B$; in both cases the package is capped at $\max(1.3B, B+300)$ lines within five iterations. After each commit the change is versioned with git, so any edit that later produces a \emph{skill\_hurt} verdict or a Task-Compiled Verifier failure can be rolled back to the previous committed state.

\paragraph{Overlapping edit conflicts.}
A surgery target and a protected segment can overlap, for example a code block that mixes a removable tutorial comment with an API call required by $\tau_w$. The Skill Iterator resolves such conflicts conservatively: protection takes precedence at the finest available granularity. The overlapping span is split at the protected boundary, only the unprotected remainder is eligible for deletion or replacement, and if a clean split is not possible the entire span is retained. This biases the framework toward under-editing rather than destroying passages required by $\tau_w$, consistent with the asymmetric cost of a harmful update.

\paragraph{Degenerate without-skill runs.}
When the without-skill trajectory $\tau_{wo}$ fails to provide a coherent reference (\S\ref{sec:method:cpa}), the Effectiveness Delta evaluators record a null incremental signal rather than a fabricated comparison, the swap protocol (which would import a verified snippet from $\tau_{wo}$) is disabled, and the iteration relies on the Task-Compiled Verifier and the single-trajectory evaluators applied to $\tau_w$. The skill can still be pruned or corrected, but supplying missing execution information from the contrast is suspended until a future iteration yields an informative pair.

\paragraph{Protocol record and reproducibility.}
All model-mediated operations in the main experiments---including paired task execution, PACE auditing, Task-Compiled Verifier generation, and skill editing---use Claude Code 2.1.158 with Claude Opus 4.8. We set the temperature to 0.1 and cap the context window at 1M tokens. Each evolution execution is limited to two hours and at most five iterations, while external benchmark evaluation is limited to one hour. Ablation variants evolve skills with GLM-5.2 and evaluate under GLM-5.2, Qwen3.8-Max, and DeepSeek-V4-Pro (\S\ref{sec:exp:ablation}). For each task, the initial audit records the matched-pair paths, categorical Effectiveness Delta, diagnosis scope, surgery targets, protected passages, locked edit mode, verifier reference, and profile fingerprint; the run record retains subsequent edit and line-count changes. These records exclude task rewards, benchmark tests, reference solutions, and benchmark verifier outputs. The released artifact will include the paper profile, controller and role prompts, the fixed template inventory, the verifier-generation specification and generation records, selection and budget rules, launch scripts, the audit-record schema, the frozen manifest, and evolved skill packages for all \numtasks{} tasks.

\paragraph{Code availability.} Code will be released upon publication.

\begin{table}[ht]
\centering
\small
\caption{Paper-mode protocol and initial-audit record. The protocol fixes the information boundary and the first routing decision without exposing benchmark verifier outputs during evolution.}
\label{tab:protocol-record}
\setlength{\tabcolsep}{4pt}
\begin{tabular}{@{}p{0.27\columnwidth}p{0.65\columnwidth}@{}}
\toprule
Component & Contract \\
\midrule
Task-Compiled Verifier & Maps explicit deterministic constraints in $T$ to checks, using $W$ only for task-referenced data; validates and locks the verifier before paired evolution; excludes skills, traces, evaluator reports, task rewards, benchmark tests, reference solutions, and benchmark verifier outputs; receives no manual task-specific edits. \\
PACE inventory & Applies the same twelve templates to every paired audit; a template without task-specific evidence emits no signal rather than being replaced or routed by an external outcome signal. \\
Initial audit & Records the paired traces, Effectiveness Delta, diagnosis scope, surgery targets, protected passages, verifier reference, profile fingerprint, and a locked Refine or Repair mode. \\
Edit policy & Permits at most five iterations; Refine and Repair add at most $0.1B$ and $0.2B$ net lines per round, respectively, with a common cap of $\max(1.3B, B+300)$. \\
\bottomrule
\end{tabular}
\end{table}

\begin{table}[t]
\centering
\normalsize
\renewcommand{\arraystretch}{1.0}
\setlength{\tabcolsep}{4pt}
\caption{Revision-method properties, grouped by the three routes described above plus the closest paired baseline (skill-creator). No outcome supervision: revision selection uses no external outcome signal; Pair: matched with/without-skill runs; Skill: revises a loadable skill package; Local edits: passage-level edits, not a wholesale rewrite.}
\label{tab:method_properties}

\begin{tabular*}{\linewidth}{@{\extracolsep{\fill}}lcccc@{}}
\toprule
Method
& No outcome supervision
& Pair
& Skill
& Local edits \\
\midrule
Trajectory memory   & \xmark & \xmark & \xmark & \xmark \\
Outcome-gated       & \xmark & \xmark & \yes   & \xmark \\
Deployment-feedback & \xmark & \xmark & \yes   & \xmark \\
skill-creator       & \xmark & \yes   & \yes   & \xmark \\
\sysname{}          & \yes   & \yes   & \yes   & \yes   \\
\bottomrule
\end{tabular*}
\end{table}

\section{Related Work}
\label{sec:appendix:related}

\paragraph{Adjacent instruction optimization and paired execution.}
DSPy, GEPA, and related prompt or context-playbook methods optimize prompts against an evaluation function or execution success~\citep{khattab2023dspycompilingdeclarativelanguage,agrawal2026gepareflectivepromptevolution,zhang2026agenticcontextengineeringevolving,yuksekgonul2024textgradautomaticdifferentiationtext}. Process-supervision methods in reasoning instead train on step-level labels or process reward models~\citep{lightman2024letsverifystepbystep,choudhury2025processrewardmodelsllm}. Anthropic's skill-creator is the closest comparison, because it also compares executions with and without a skill, but practitioner-supplied test assertions still decide whether to retain an edit~\citep{anthropic2025skills}. In contrast, \sysname{} uses the behavioral differences between the two executions to select skill passages for revision and a Task-Compiled Verifier to reject violations of explicit task constraints, without task rewards, held-out validation results, or practitioner-supplied test assertions.

\section{Evolution Loop Algorithm}
\label{sec:appendix:algo}

\begin{algorithm}[H]
\caption{\sysname{} Evolution Loop}
\label{alg:loop}
\begin{algorithmic}[1]
\REQUIRE Task description $T$, workspace $W$, initial skill $S_0$
\ENSURE Evolved skill $S^*$
\STATE $\mathcal{I} \leftarrow \textsc{TaskInterpreter}(T)$ \hfill\COMMENT{explicit constraints from the task}
\STATE $V \leftarrow \textsc{CompileTaskVerifier}(\mathcal{I}, W)$ \hfill\COMMENT{$W$ grounds task-referenced data}
\STATE $(\tau_w, \tau_{wo}) \leftarrow \textsc{PairedExecution}(T, W, S_0)$
\STATE $D_0 \leftarrow \textsc{PACE}(\tau_w, \tau_{wo}, S_0)$ \hfill\COMMENT{first paired diagnosis}
\STATE $\textit{mode} \leftarrow \textsc{SelectMode}(\textsc{EffectivenessDelta}(D_0), \textsc{DiagnosisScope}(D_0))$ \hfill\COMMENT{lock Refine or Repair}
\STATE $S \leftarrow S_0$;\; $S_{\mathrm{acc}} \leftarrow S_0$;\; $H \leftarrow \varnothing$
\FOR{$t = 1, \ldots, 5$}
    \IF{$t > 1$}
        \STATE $\tau_w, \tau_{wo} \leftarrow \textsc{PairedExecution}(T, W, S)$
        \STATE $D_t \leftarrow \textsc{PACE}(\tau_w, \tau_{wo}, S)$
    \ELSE
        \STATE $D_t \leftarrow D_0$
    \ENDIF
    \STATE $\textit{verdict} \leftarrow \textsc{Aggregate}(D_t)$ \hfill\COMMENT{hurt has veto; mode unchanged}
    \STATE $H \leftarrow H \,\|\, [\textit{verdict}]$ \hfill\COMMENT{append to verdict history}
    \IF{$S \neq S_{\mathrm{acc}}$ \AND (\textit{verdict} = \textit{hurt} \OR $V(\tau_w)$ fails)}
        \STATE $S \leftarrow S_{\mathrm{acc}}$ \hfill\COMMENT{reject the edited candidate}
    \ELSE
        \IF{$V(\tau_w)$ passes}
            \STATE $S_{\mathrm{acc}} \leftarrow S$
        \ENDIF
        \IF{$t<5$ \AND \textsc{HasActionablePassages}$(D_t, \textit{mode})$}
            \STATE $S \leftarrow \textsc{SkillIterator}(S, \tau_w, \tau_{wo}, \textit{mode})$
        \ENDIF
    \ENDIF
    \IF{$S = S_{\mathrm{acc}}$ \AND \textsc{Converged}$(D_t, V(\tau_w), H)$} \STATE \textbf{break} \ENDIF
\ENDFOR
\RETURN $S_{\mathrm{acc}}$
\end{algorithmic}
\end{algorithm}

\textsc{Aggregate} follows the \textsc{PACE} precedence rule in \S\ref{sec:method:pace}; the Task-Compiled Verifier remains a separate acceptance gate. \textsc{Converged} requires the verifier to pass, no \emph{hurt} verdict in the last two iterations, and no actionable passages.

The pair computed before the loop is the first paired audit of $S_0$ and is reused in iteration $t=1$. Its Effectiveness Delta and diagnosis scope select the locked mode; later iterations never reroute Refine and Repair. A \emph{hurt} verdict or verifier failure rejects only a candidate produced by \textsc{SkillIterator} and restores $S_{\mathrm{acc}}$. Otherwise, actionable passages identified by \textsc{PACE} trigger an edit under the locked mode, including when the first audit selects Repair for an initially harmful or inert skill. The edited package is evaluated by the with-skill execution of the next iteration. The loop terminates early when no actionable passages remain, recent updates are not harmful, and the verifier passes.

\begin{table}[t]
\centering
\small
\caption{Which updates Refine and Repair permit. Both use paired trajectories and the Task-Compiled Verifier.}
\label{tab:dual}
\begin{tabular}{@{}l p{0.24\textwidth} p{0.38\textwidth}@{}}
\toprule
Dimension & Refine & Repair \\
\midrule
Current workflow & Kept & May be replaced \\
Default action & Delete, clarify, or correct locally & Replace the prescribed workflow \\
Protected passages & Marked by \textsc{PACE} to keep; not edited & All passages are evaluated by \textsc{PACE}; diagnosed protected passages are retained \\
Harmful content & Reject the update & Locate via PACE, replace or delete \\
Missing information & Local details only (unit, header) & Added after harmful passages are removed \\
\bottomrule
\end{tabular}
\end{table}

\section{PACE Evaluator Inventory}
\label{sec:appendix:pace}

PACE is built from a fixed inventory of twelve evaluator templates spanning the four dimensions introduced in \S\ref{sec:method:pace}. Table~\ref{tab:pace} gives one representative evaluator per dimension, and Table~\ref{tab:pace_full} gives the complete inventory. The same evaluator inventory is used for every paired audit; evaluators without relevant evidence return no signal or mark the criterion as not applicable. Each evaluator reads both $\tau_w$ and $\tau_{wo}$, compares tool calls, decisions, and artifacts, and emits passage-level \texttt{action\_signals} and \texttt{protected\_hints} as described in \S\ref{sec:method:pace}.

\paragraph{Evaluator instantiation and information boundary.}
PACE combines this fixed inventory with task-specific grounding. Before evolution, the evaluator generator binds every immutable template to the explicit requirements in $T$, the declared workspace data in $W$, and the passage structure of the current skill. During each paired audit, every evaluator jointly examines $\tau_w$ and $\tau_{wo}$, traces observed behavioral differences to specific skill passages, and emits structured \texttt{action\_signals} and \texttt{protected\_hints}. A protocol-level validator enforces outcome-supervision-free evolution by rejecting any evaluator payload containing task rewards, benchmark tests, benchmark-verifier outputs, reference solutions, or historical comparison reports. Paper-formatted versions of \texttt{eval-output-evidence-check} and \texttt{eval-skill-incremental-value} appear below; the supplementary artifact provides all twelve templates and their binding procedure.

\begin{table}[H]
\centering
\small
\caption{PACE evaluation dimensions, with one representative evaluator per dimension. Each evaluator compares $\tau_w$ and $\tau_{wo}$ along one dimension and outputs passage-level diagnostic signals.}
\label{tab:pace}
\begin{tabular}{@{}p{2.5cm}p{3.7cm}p{6.9cm}@{}}
\toprule
Dimension & Representative evaluator & What is assessed \\
\midrule
Process Adherence & eval-skill-procedure-adherence & Does the agent follow the skill's \\
& & prescribed steps and tool-use patterns? \\
\addlinespace
Artifact Evidence & eval-output-evidence-check & Do the output files meet explicit \\
& & task requirements and deterministic workspace checks? \\
\addlinespace
Consistency & eval-skill-task-alignment & Does the skill's prescribed workflow match \\
& & the task's data schema and output format? \\
\addlinespace
Effectiveness Delta & eval-skill-incremental-value & Does $\tau_w$ improve on $\tau_{wo}$ in \\
& & calls, decisions, or artifacts? \\
\bottomrule
\end{tabular}
\end{table}

\begin{table}[H]
\centering
\small
\caption{Complete fixed PACE evaluator inventory (12 templates across four dimensions). Every template is applied to each paired audit; a task without relevant evidence yields no signal or a not-applicable record.}
\label{tab:pace_full}
\setlength{\tabcolsep}{4pt}
\begin{tabular}{@{}p{2.4cm} p{3.8cm} p{7.1cm}@{}}
\toprule
Dimension & Evaluator & What it checks \\
\midrule
Process Adherence
  & eval-skill-procedure-adherence & Does the agent execute the skill's prescribed steps in order, respecting preconditions and prohibited actions? \\
  & eval-skill-coverage & What fraction of the agent's key actions are guided by explicit skill instructions vs.\ self-inferred? \\
  & eval-tool-use-rationality & Are tool choice, call timing, parameter construction, and use of returned values reasonable? \\
  & eval-error-robustness & Does the agent handle errors and exceptions sensibly rather than failing silently or retrying without bound? \\
\addlinespace
Artifact Evidence
  & eval-output-evidence-check & Three-way comparison of observable tokens across the with-skill artifact, the without-skill artifact, and workspace data, with skill attribution. \\
  & eval-format-compliance & Does the output conform to the format contract (structure, field completeness, types, enum validity)? \\
\addlinespace
Consistency
  & eval-skill-task-alignment & Do the skill's instructions contradict the task requirements (over-specification, workflow conflicts)? \\
  & eval-skill-data-consistency & Do the skill's data assumptions (file/column/field names, parameter values) match the actual data? \\
  & eval-method-adherence & Does the agent use the specific method, algorithm, or tool the task mandates? \\
  & eval-safety-compliance & Does execution violate a hard constraint whose breach is not recoverable within the task (unsafe paths, credentials, data protection, or destruction of content the task requires to be preserved)? Highest priority; vetoes on a critical violation. \\
\addlinespace
Effectiveness Delta
  & eval-skill-incremental-value & Does the with-skill run provide a genuine procedural gain over the without-skill run, and does the gain come from the skill rather than native model ability? \\
  & eval-skill-portfolio-quality & Audits skill-portfolio health: redundancy, document bloat (line/file inflation), and unilateral growth of imperative instructions. \\
\bottomrule
\end{tabular}
\end{table}

\paragraph{Complete representative evaluator prompt: \texttt{eval-output-evidence-check}.}
Among the twelve fixed evaluators, \texttt{eval-output-evidence-check} supplies the artifact-level evidence used by PACE. It performs a three-way comparison among declared workspace data, artifacts produced with and without the skill, and the skill passages associated with the observed behavior. When task-grounded artifact evidence resolves the comparison, this evaluator is decisive in PACE aggregation. The paper-formatted prompt below preserves the complete decision procedure and output contract; symbolic paths are replaced with task-specific values during binding.

\begingroup
\small
\setlength{\parindent}{0pt}
\setlength{\parskip}{2pt}
\hrule
\vspace{4pt}
\textbf{Role and objective.}
You are the Artifact Evidence evaluator in PACE. Determine whether the artifacts produced by the with-skill and without-skill executions satisfy the file-level and content-level requirements grounded in the task and workspace. Compare both conditions, identify skill-induced gains or regressions, and connect every actionable finding to the responsible skill passage.

\textbf{Bound inputs.}
The evaluator receives: (i) the task description $T$; (ii) task-grounded \texttt{requirements}, \texttt{observable\_tokens}, expected output files, hard constraints, and a profile of declared workspace data $W$; (iii) the current skill $S$; (iv) the matched trajectories $\tau_w$ and $\tau_{wo}$; (v) their respective artifact directories; and (vi) the destination for the structured report. Task binding replaces these fields without changing the checks or decision rules below.

\textbf{Evidence policy.}
Use only observable facts from $T$, $W$, $S$, the paired trajectories, and their generated artifacts. Inspect files with controlled reads, exact search, and format-aware parsers. Every finding must include a reproducible file excerpt, parsed value, or trace record; never invent an event or artifact. Do not access benchmark tests, benchmark-verifier outputs, scoring scripts, reference solutions, task rewards, or historical comparison reports. Task requirements are grounded in $T$ and $W$; $S$ is used to attribute observed behavior to an editable passage.

\textbf{Evaluation procedure.}
\begin{enumerate}[leftmargin=1.6em,itemsep=2pt,topsep=2pt]
    \item \textbf{Artifact existence and completeness.} For every expected output, check that the file exists, is nonempty, and can be parsed when its required format is structured, such as JSON, CSV, or XLSX.
    \item \textbf{Three-way observable-evidence comparison.} For every task-grounded observable token, locate its source value in $W$, test its presence or required transformation in each output, and locate the passage in $S$ associated with that behavior. A source value present in $W$ and correctly propagated to an artifact is a pass; a source value present but missing from the artifact is a failure; an absent source value makes the check not applicable.
    \item \textbf{Numerical and structural consistency.} Check explicit cross-field relations stated in $T$, including sums, percentages, identifiers, schemas, and preservation requirements. When the declared task structure supports an additional consistency comparison, record it as warning-level evidence rather than a hard requirement.
    \item \textbf{With-skill versus without-skill comparison.} Apply the same checks independently to both artifact sets. Evidence present only in the with-skill artifact indicates a skill-enabled gain. Evidence present in the without-skill artifact but missing or degraded in the with-skill artifact is a candidate \texttt{skill\_introduced\_regression}.
    \item \textbf{Skill attribution.} For every failure or regression, match the affected requirement and trace behavior to the responsible skill file, section heading, and verbatim passage. The passage location must be specific enough for the Skill Iterator to edit without rewriting unrelated content.
    \item \textbf{Regression localization.} For a high-confidence regression, connect the affected evidence to the most recent relevant skill change and report the changed passage, affected token or invariant, and observed mechanism. Prefer restoring the last supported passage when the evidence identifies a regression.
\end{enumerate}

\textbf{Required output.}
Emit one structured record with the following fields:
\begin{itemize}[leftmargin=1.6em,itemsep=1pt,topsep=2pt]
    \item \texttt{evaluator\_name}, fixed to \texttt{eval-output-evidence-check}; \texttt{dimension}, fixed to \texttt{artifact\_evidence}; and \texttt{agent\_mode}, identifying the with-skill, without-skill, or paired summary record.
    \item \texttt{findings[]}, where each entry contains the requirement identifier, checked token or invariant, source presence, artifact presence, result, confidence, responsible skill passage, and concrete evidence.
    \item \texttt{regression\_findings[]}, where each entry contains the regressed token or invariant, the \texttt{skill\_introduced\_regression} verdict, confidence, responsible skill passage, and the associated skill change.
    \item \texttt{completeness\_score}, \texttt{token\_pass\_rate}, overall result, severity, attribution, and a passage-specific suggestion.
    \item \texttt{action\_signals[]}. For each high-confidence skill-induced regression, emit \texttt{action=investigate\_regression}, the exact skill file and passage, confidence, and an evidence-grounded rationale. Otherwise emit an empty list.
    \item \texttt{protected\_hints[]}, emitted as an empty list by this evaluator; preservation signals are supplied by the evaluators responsible for confirmed procedural value.
\end{itemize}

\textbf{Aggregation contract.}
When task-grounded artifact checks resolve the comparison, return the artifact verdict as decisive evidence to PACE. Every signal must remain traceable from a task requirement or declared workspace fact, through the paired artifacts, to the cited skill passage.
\vspace{4pt}
\hrule
\endgroup

\paragraph{Complete representative evaluator prompt: \texttt{eval-skill-incremental-value}.}
This evaluator operationalizes the Effectiveness Delta dimension of \textsc{PACE}. It compares matched executions, distinguishes skill-grounded behavioral differences from native-model variation, and links each supported difference to the responsible skill passage. Its comparison record supplies the incremental-value judgment used by the initial routing rule. The paper-formatted prompt below preserves the decision procedure and output contract while omitting serialization details.

\begingroup
\small
\setlength{\parindent}{0pt}
\setlength{\parskip}{2pt}
\hrule
\vspace{4pt}
\textbf{Role and objective.}
You are the Effectiveness Delta evaluator in \textsc{PACE}. Determine whether loading the current skill provides a genuine procedural contribution beyond the frozen agent's behavior without that skill. Compare the matched trajectories, identify the passages associated with supported behavioral differences, and separate skill-grounded effects from native-model variation.

\textbf{Bound inputs.}
The evaluator receives: (i) the task description $T$ and declared workspace data $W$; (ii) the current skill $S$ and its passage structure; (iii) the matched trajectories $\tau_w$ and $\tau_{wo}$; (iv) their generated artifacts; and (v) the destination for the structured records. Task binding may identify expected divergence points, but it does not change the comparison rules below.

\textbf{Evidence policy.}
Use only observable calls, decisions, tool results, and artifacts from the matched executions. Do not access task rewards, benchmark tests, benchmark-verifier outputs, reference solutions, or historical comparison reports. Attribute a difference to the skill only when the observed behavior matches a concrete operation stated in a specific skill passage. Otherwise record the source as native-model behavior or unclear; such differences do not authorize a skill edit.

\textbf{Evaluation procedure.}
\begin{enumerate}[leftmargin=1.6em,itemsep=2pt,topsep=2pt]
    \item \textbf{Map passages to behavioral signatures.} Read the skill and associate each procedural passage with observable signatures such as tool choice, call order, arguments, paths, or artifact transformations.
    \item \textbf{Identify paired divergences.} Compare $\tau_w$ and $\tau_{wo}$ at corresponding task steps, using tool calls, tool results, decisions, and artifacts to locate where their behavior diverges.
    \item \textbf{Determine the source.} Mark a divergence as \texttt{skill\_knowledge} only when the with-skill behavior matches a cited skill passage and the without-skill run lacks the corresponding behavior. Mark an ungrounded difference as \texttt{model\_native}, and use \texttt{unclear} when the available trace cannot distinguish the source.
    \item \textbf{Determine incremental value.} Return \texttt{positive} when a skill-grounded divergence improves an observable task step or artifact; \texttt{negative} when it introduces an error, detour, or degraded artifact; \texttt{zero} when both runs behave equivalently; and \texttt{zero\_missing\_knowledge} when the skill omits needed guidance that the agent supplies from native capability. If the traces are incomplete, return \texttt{unclear} and do not force a conclusion.
    \item \textbf{Inspect unresolved stalls.} For a zero or unclear comparison, locate the last productive tool call on each side and record whether either execution makes materially deeper progress before repeated errors, abandonment, or prolonged absence of productive tool use.
    \item \textbf{Produce passage-level signals.} A positive, skill-grounded difference protects the cited passage. A negative, skill-grounded difference marks that passage for deletion or replacement and, when available, records the corresponding verbatim tool-use fragment from $\tau_{wo}$. Zero, native-model, and unclear differences do not create an edit target.
\end{enumerate}

\textbf{Required output.}
Emit one record for each execution and one comparison record. Each record contains \texttt{evaluator\_name}, \texttt{dimension}, \texttt{agent\_mode}, a factual \texttt{phenomenon}, traceable \texttt{evidence}, and any identified error. The comparison record additionally contains \texttt{incremental\_value}, \texttt{incremental\_source}, optional stall-depth fields, and \texttt{without\_verified\_snippets[]} for a supported negative difference. Emit \texttt{action\_signals[]} only for negative differences attributed to skill knowledge, and emit \texttt{protected\_hints[]} only for positive differences attributed to skill knowledge. Every signal must identify the skill file and quote the responsible passage verbatim.

\textbf{Aggregation contract.}
PACE accepts exactly one comparison record from this evaluator. A positive comparison supplies positive Effectiveness Delta evidence, a negative comparison supplies negative evidence, and \texttt{zero} or \texttt{zero\_missing\_knowledge} supplies near-zero evidence. The initial routing rule combines this judgment with diagnosis scope to select Refine or Repair.
An unclear comparison does not authorize an edit.
\vspace{4pt}
\hrule
\endgroup

\section{Supplementary Evidence and Case Studies}
\label{sec:appendix:cases}

Each case below illustrates one component of the method, stated in bold. \bad{Red} marks the defective side of a contrast: a harmful instruction, or an observation that fails a constraint; \ok{blue} marks the correct side: the without-skill reference, a corrected fragment, or an observation that satisfies the constraint. Cases whose claim depends on the loop include an iteration table. Figure~\ref{fig:case-verifier} shows the constraint-check rollback and the later Repair steps on \texttt{data-to-d3}. Figure~\ref{fig:case-pair} compares the two pairing arms on \texttt{grid-dispatch-operator}. Cases~A and~B are the examples in \S\ref{sec:exp:analysis}. We will release the evolved skill packages with the implementation.

\begin{table}[h]
\centering
\small
\caption{Four Opus case studies. Each row illustrates one mechanism: A,~Refine keeps a working workflow; B,~the framework rejects a candidate whose with-skill execution violates an explicit constraint; C,~an inert round does not switch the fixed mode; D,~a hurt verdict vetoes the update without recomputing the mode.}
\label{tab:casestudy}
\setlength{\tabcolsep}{4pt}
\begin{tabular}{@{}l l r r p{3.1cm} r@{}}
\toprule
Task & Mode & Lines & $\Delta$ (\%) & Paired contrast & \shortstack{Opus\\task reward} \\
\midrule
\texttt{software-dependency-audit} & Refine & 935$\to$610 & $-$35 & iter~2: $\tau_w$ PASS / $\tau_{wo}$ FAIL & 1.0 \\
\texttt{data-to-d3} & Repair & 188$\to$50 & $-$73 & iter~1 revert; iter~3: $\tau_w$ PASS / $\tau_{wo}$ FAIL & 1.0 \\
\texttt{lab-unit-harmonization} & Refine & 408$\to$259 & $-$37 & iter~2: $\tau_w$ PASS 8/8 / $\tau_{wo}$ FAIL 7/8 & 1.0 \\
\texttt{exceltable-in-ppt} & Repair & 1825$\to$1773 & $-$3 & iter~2: $\tau_w$ FAIL (formula hurt) $\to$ rollback; iter~4: converge & 1.0 \\
\bottomrule
\end{tabular}
\end{table}

\subsection{Case A: \texttt{software-dependency-audit} (Refine, primary)}
\label{sec:appendix:case-a}

\paragraph{Mode.}
\textbf{Refine keeps a working workflow and deletes passages that $\tau_w$ does not use.}

Offline Trivy scan of \texttt{package-lock.json} into a CSV with a fixed header. Opus evolved reward $1.0$.

\begin{table}[h]
\centering
\small
\setlength{\tabcolsep}{4pt}
\begin{tabular}{@{}c l l l r@{}}
\toprule
Iter & $\tau_w$ & $\tau_{wo}$ & Decision & Lines \\
\midrule
1 & PASS & PASS & delete unused tutorial; replace cache path; keep offline flags & $935\to610$ \\
2 & PASS & FAIL & \emph{skill\_helped}; stop & 610 \\
\bottomrule
\end{tabular}
\end{table}

\paragraph{Contrast.}
\bad{Hardcoded \texttt{./trivy-cache}} vs.\ \ok{\texttt{--cache-dir} with a parameter}. Protected segments keep \texttt{--skip-db-update} and \texttt{--offline-scan}. Without-skill omits those flags. The prescribed scan workflow is not replaced.

\subsection{Case B: \texttt{data-to-d3} (Repair, primary)}
\label{sec:appendix:case-b}

\paragraph{Mode.}
\textbf{The framework rejects a candidate whose with-skill execution violates a constraint already satisfied without the skill, and restores the previously accepted skill.}

Build \texttt{/root/output/index.html} with D3 bubbles and a 50-row table whose column strings must match the task verbatim. The compiled verifier instantiates 13 checks, including the verbatim column-string requirement. At iteration~1, Title Case headers violate that check and yield a 12/13 result; the corrected artifact later passes all 13 checks. Opus $0.0\to1.0$. Figure~\ref{fig:case-verifier} is the recorded process; the table lists every archived round.

\begin{figure}[H]
\centering
\resizebox{\textwidth}{!}{%
\begin{tikzpicture}[
  font=\scriptsize,
  box/.style={draw, rounded corners=2pt, align=center, inner sep=3pt, text width=2.42cm, minimum height=1.62cm, line width=0.7pt},
  arr/.style={-{Latex[length=4.5pt]}, thick, black!55},
  dasharr/.style={-{Latex[length=4.5pt]}, thick, dashed, black!45},
]
\node[box, fill=black!4] (s0) {\textbf{$S_0$}\\188 lines\\Repair locked};
\node[box, draw=saRed, fill=saRed!8, right=0.22cm of s0] (i1) {\textbf{Iter.\ 1}\\\bad{$\tau_w$ 12/13 FAIL}\\\ok{$\tau_{wo}$ 13/13 PASS}};
\node[box, draw=saRed, fill=saRed!8, right=0.22cm of i1] (rev) {\textbf{Revert}\\\bad{back to $S_0$}\\iterator not called};
\node[box, draw=saBlue, fill=saBlue!8, right=0.22cm of rev] (i3) {\textbf{Iter.\ 3}\\48 lines\\\ok{$\tau_w$ 13/13 PASS}\\\bad{$\tau_{wo}$ 12/13 FAIL}};
\node[box, draw=saBlue, fill=saBlue!8, right=0.22cm of i3] (i5) {\textbf{Iter.\ 5}\\50 lines\\\ok{13/13 PASS}\\verbatim label};
\draw[arr] (s0) -- (i1);
\draw[arr] (i1) -- (rev);
\draw[dasharr] (rev) -- (i3);
\draw[arr] (i3) -- (i5);
\end{tikzpicture}%
}
\caption{\texttt{data-to-d3}: the framework rejects a candidate whose with-skill execution fails a constraint already met without the skill, then continues Repair under the same mode. Dashed: iteration~2 is not in the archived excerpt.}
\label{fig:case-verifier}
\end{figure}

\begin{table}[h]
\centering
\small
\setlength{\tabcolsep}{3.5pt}
\begin{tabular}{@{}c l l l r@{}}
\toprule
Iter & $\tau_w$ & $\tau_{wo}$ & Decision & Lines \\
\midrule
1 & 12/13 FAIL (columns) & 13/13 PASS & revert to $S_0$; no iterator & 188 \\
2 & --- & --- & not in archived excerpt & --- \\
3 & 13/13 PASS & 12/13 FAIL & \emph{skill\_helped} & 48 \\
4 & 12/13 FAIL (Title Case) & unstable & journal: agent styling, not skill & 48 \\
5 & 13/13 PASS & --- & verbatim-label reminder; converge & 50 \\
\bottomrule
\end{tabular}
\end{table}

\paragraph{Contrast.}
\bad{With-skill Title Case headers} (\texttt{Ticker}, \texttt{Company Name}) vs.\ \ok{without-skill verbatim columns} (\texttt{Ticker symbol}, \texttt{Full company name}). Iteration~1 rolls back on that contrast. Later Repair keeps tooltip and click snippets and adds a verbatim-label reminder.

\subsection{Case C: \texttt{lab-unit-harmonization} (Refine)}
\label{sec:appendix:case-c}

\paragraph{Mode.}
\textbf{An inert round does not switch the locked Refine mode; the paired contrast can appear later.}

SI unit harmonization of a clinical CSV. Opus evolved reward $1.0$. The first paired audit judges the with-skill run to provide positive incremental value, so the locked mode is Refine; the \emph{skill\_inert} verdict in iteration~1 records that no actionable surgery target was found in that round, not that the initial edit-mode choice changes.

\begin{table}[h]
\centering
\small
\setlength{\tabcolsep}{4pt}
\begin{tabular}{@{}c l l l r@{}}
\toprule
Iter & $\tau_w$ & $\tau_{wo}$ & Decision & Lines \\
\midrule
1 & PASS 8/8 & PASS 8/8 & \emph{skill\_inert}; no edit; mode stays Refine & 408 \\
2 & PASS 8/8 & FAIL 7/8 & protect unit table; delete unused text; \emph{skill\_helped} & 259 \\
\bottomrule
\end{tabular}
\end{table}

\paragraph{Contrast.}
Iteration~1 has no failure that requires Repair. Iteration~2: \ok{$\tau_w$ converts a legacy \texttt{mg/dL} field} vs.\ \bad{$\tau_{wo}$ omits that mapping}. The unit-mapping table stays; surrounding unused passages are deleted.

\subsection{Case D: \texttt{exceltable-in-ppt} (Repair)}
\label{sec:appendix:case-d}

\paragraph{Mode.}
\textbf{A hurt verdict vetoes the current update and does not recompute the locked mode. This case is not evidence for the Task-Compiled Verifier.}

Embed an Excel sheet into PowerPoint while keeping six cell formulas. Curated and evolved rewards are both $1.0$. Here the first paired audit finds no positive incremental value, so the locked mode is Repair, even though both runs of iteration~1 pass the verifier as in Case~C.

\begin{table}[h]
\centering
\small
\setlength{\tabcolsep}{3.5pt}
\begin{tabular}{@{}c l l l r@{}}
\toprule
Iter & $\tau_w$ & $\tau_{wo}$ & Decision & Lines \\
\midrule
1 & PASS 6/6 & PASS 6/6 & \emph{skill\_inert}; no edit & 1825 \\
2 & FAIL 0/6 & PASS 6/6 & \emph{skill\_hurt}; rollback; mode stays Repair & 1825 \\
3 & PASS 6/6 & PASS 6/6 & delete \texttt{recalc.py} step; \emph{skill\_helped} & 1773 \\
4 & PASS 6/6 & PASS 6/6 & verification; converge & 1773 \\
\bottomrule
\end{tabular}
\end{table}

\paragraph{Contrast.}
\bad{\texttt{MANDATORY IF USING FORMULAS: Use the recalc.py script}} zeroes formulas on this task ($0/6$). \ok{The hurt veto rolls back and leaves the mode Repair.} A later Repair iteration deletes that step ($-52$ lines). The formula check that fires is \emph{eval-safety-compliance}, not the Task-Compiled Verifier.

\subsection{Case E: \texttt{paper-anonymizer} (Repair; passage-level diagnosis)}
\label{sec:appendix:case-pace}

\paragraph{Mode.}
\textbf{\textsc{PACE} names the passage to edit.} Repair; the task's reward rises from $0$ to $1$ after evolution (no-skill $0$, initial skill $0$, evolved skill $1$; Table~\ref{tab:pertask}).

Anonymize three academic PDFs for blind review (\texttt{/root/paper\{1-3\}.pdf} $\to$ \texttt{/root/redacted/}): redact author names, affiliations, arXiv identifiers, and accepted venues while preserving the References section. The task specification requires that at least $60\%$ ($0.6$) of the source text be retained.

\paragraph{Contrast.}
Initial skill (\texttt{academic-pdf-redaction/SKILL.md}). Its retention guard is both self-contradictory and stricter than the task, in prose and in code:
\begin{quote}\small
\bad{\texttt{VERIFY output - Check that 80\%+ of original text remains}}\\
\bad{\texttt{if redc\_len < orig\_len * 0.7: raise ValueError(...)}}
\end{quote}
The prose ($80\%$) and the code ($0.7$) disagree, and both exceed the task's $0.6$ floor, so the guard rejects a faithful redaction as a failure. \textsc{PACE} marks the guard as edit targets ST-002/ST-003 and separately unwraps a generic \texttt{redact\_with\_pymupdf(...)} function wrapper (ST-001), protecting the remaining sections. After the edit the guard reads:
\begin{quote}\small
\ok{\texttt{Check that 60\%+ of original text remains}}\\
\ok{\texttt{if redc\_len < orig\_len * 0.6: raise ValueError(...)}}
\end{quote}
A later round confirms that the diagnosed issues have been resolved. During the initial audit, the diagnosis authorizes removal of off-domain files bundled with the skill ($1223 \to 112$ lines); the in-domain redaction module is not rewritten.

\subsection{Case F: \texttt{grid-dispatch-operator} (Refine; pairing contrast)}
\label{sec:appendix:case-pair}

\paragraph{Mode.}
\textbf{A without-skill run is required for the update signal.} Two with-skill runs can treat an execution accident as skill harm and edit a working workflow. GLM pairing run (\S\ref{sec:exp:ablation}), not the Opus \numtasks{}-task table. Figure~\ref{fig:case-pair} places the two arms side by side.

\begin{figure}[H]
\centering
\begin{tikzpicture}[
  font=\scriptsize,
  box/.style={draw, rounded corners=2pt, align=center, inner sep=4pt, text width=0.38\textwidth, minimum height=1.55cm, line width=0.7pt},
  arr/.style={-{Latex[length=4.5pt]}, thick, black!55},
]
\begin{scope}
  \node[font=\small\bfseries, text=saBlue] (At) {Arm A: with / without};
  \node[box, draw=saBlue, fill=saBlue!8, below=0.18cm of At] (A1) {\textbf{Iter.\ 1}\\both checks pass; with-skill cost optimal\\\ok{\emph{skill\_helped}; surgery $=0$; no edit (538 lines)}};
  \node[box, draw=saBlue, fill=saBlue!8, below=0.42cm of A1] (A2) {\textbf{Iter.\ 2}\\$\tau_w$ still optimal; $\tau_{wo}$ fails load balance\\Task reward \textbf{1.0}};
  \draw[arr] (A1) -- (A2);
\end{scope}
\begin{scope}[xshift=0.52\textwidth]
  \node[font=\small\bfseries, text=saRed] (Bt) {Arm B: two with-skill};
  \node[box, draw=saRed, fill=saRed!8, below=0.18cm of Bt] (B1) {\textbf{Iter.\ 1}\\no $\tau_{wo}$; $0.46\%$ reported loading vs.\ \bad{$331\%$ DC overload}\\\bad{\emph{skill\_hurt}; edits power balance and solver ($536\to560$)}};
  \node[box, draw=saRed, fill=saRed!8, below=0.42cm of B1] (B2) {\textbf{Iter.\ 2}\\structural checks still pass; stop\\Task reward \textbf{0.0}};
  \draw[arr] (B1) -- (B2);
\end{scope}
\end{tikzpicture}
\caption{\texttt{grid-dispatch-operator} in the GLM pairing run. Arm~A sees that unloading the skill is worse and leaves the working workflow unchanged. Arm~B has no without-skill reference, treats a silent infeasibility as skill harm, and rewrites passages that still pass structural checks.}
\label{fig:case-pair}
\end{figure}

\paragraph{Contrast.}
Arm~A: with-skill cost $\$2{,}965{,}637.91$/hr (optimal) vs.\ without-skill $\$3{,}086{,}631.85$; \ok{unloading the skill is worse}, so the package is not edited. Arm~B: \bad{a $0.46\%$ reported loading against a $331\%$ DC overload} is treated as skill harm. The solver rewrite still passes structural checks; task reward is $0.0$ (cost error $\approx 4\%$; tolerance $0.01\%$).

\subsection{Structural Analysis Details}
\label{sec:appendix:length}

\begin{figure}[H]
\centering
\includegraphics[width=\textwidth]{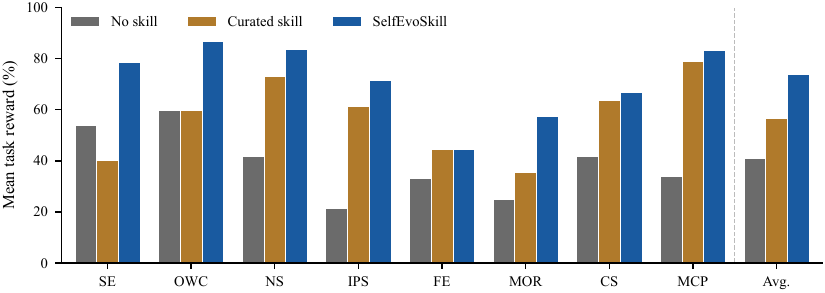}
\caption{The evolved skill leads the curated skill in seven of eight domains. Mean task reward (\%) by domain from Table~\ref{tab:main}. Domain codes match Appendix~\ref{sec:appendix:tasklevel}. The dashed line separates the eight domain means from the overall mean across all \numtasks{} tasks.}
\label{fig:mainbars}
\end{figure}

\paragraph{Domain-level reward profile.}
Figure~\ref{fig:mainbars} complements the task-level analysis in \S\ref{sec:exp:analysis} by displaying the mean reward profile behind Table~\ref{tab:main}. The evolved skill leads the curated skill in seven of eight domains; the variation motivates examining which forms of skill knowledge leave diagnostic traces in trajectories and artifacts.

\begin{table}[h]
\centering
\small
\caption{Skill line-count change ($\Delta=\text{evolved}-\text{initial}$, all \texttt{.md} files) by edit scope. Refine's long deletion tail reflects that it may not replace the prescribed workflow; Repair makes more targeted edits. Throughout this appendix, \emph{pass} denotes an evolved task reward $\ge 0.5$. Each entry pairs task reward with the line-count change of the same single evolved skill: the run that completed without an infrastructure abort. We do not use the best-of-two reward from \S\ref{sec:exp:setup}, because it does not correspond to a single skill. Figure~\ref{fig:skill_delta} uses the same convention.}
\label{tab:lengthdyn}
\setlength{\tabcolsep}{6pt}
\begin{tabular}{@{}l r r r r r@{}}
\toprule
Edit scope & $N$ & Pass / Fail & Median $\Delta$ (pass) & Mean $\Delta$ (all) & Min $\Delta$ \\
\midrule
Refine & 43 & 39 / 4  & $+19$ & $-260$ & $-7{,}375$ \\
Repair & 46 & 28 / 18 & $+16$ & $-165$ & $-2{,}867$ \\
\bottomrule
\end{tabular}
\end{table}

\begin{figure*}[t]
\centering
\begin{subfigure}[t]{0.48\textwidth}
\centering
\includegraphics[width=\textwidth]{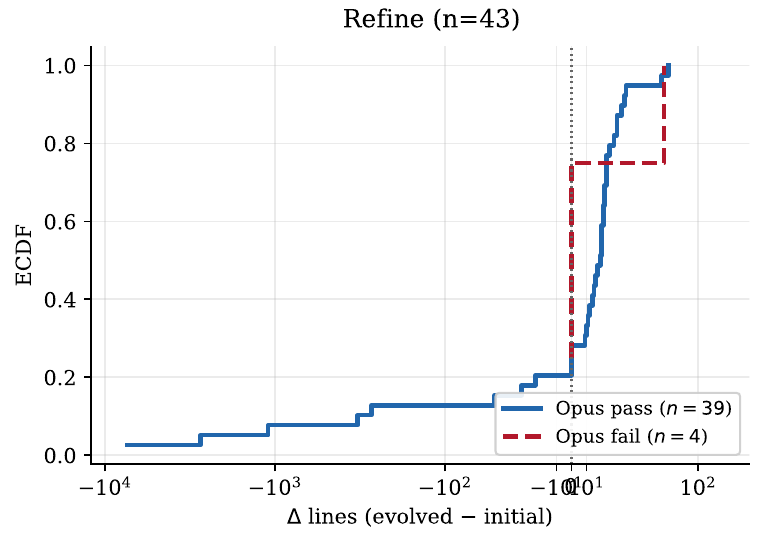}
\caption{Refine ($n=43$): pass $n=39$, fail $n=4$.}
\label{fig:skill_delta_refine}
\end{subfigure}
\hfill
\begin{subfigure}[t]{0.48\textwidth}
\centering
\includegraphics[width=\textwidth]{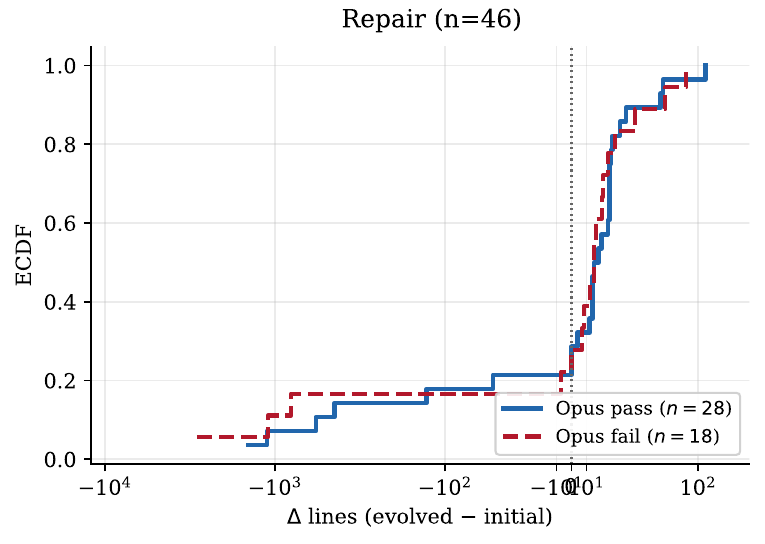}
\caption{Repair ($n=46$): pass $n=28$, fail $n=18$.}
\label{fig:skill_delta_repair}
\end{subfigure}
\caption{Empirical CDFs of skill line-count change ($\Delta = \text{evolved} - \text{initial}$) for all \numtasks{} tasks, stratified by edit scope and Opus pass/fail. Solid: pass; dashed: fail. Dotted line marks $\Delta=0$; symmetric log scale (linear for $|\Delta|\le 50$).}
\label{fig:skill_delta}
\end{figure*}

Table~\ref{tab:lengthdyn} summarizes the line-count dynamics by edit scope: Refine has a median of $+19$ lines on passing tasks alongside a deletion tail reaching $-7{,}375$, while Repair makes more targeted swaps with fewer extreme deletions. Figure~\ref{fig:skill_delta} gives the full per-task distribution of skill line-count change, as empirical CDFs stratified by edit scope and post-evolution Opus pass/fail. The horizontal axis uses a symmetric log scale so that large deletions (e.g., \texttt{flink-query}, $-7{,}375$ lines) remain visible without compressing the central mass near zero.

\section{Task Manifest, Per-Task Results, and Label Statistics}
\label{sec:appendix:tasklevel}

\paragraph{Per-label statistics.}
Table~\ref{tab:labelstats} reports the average evolved reward for every knowledge-type and task-type label. Labels are not mutually exclusive: per-label counts are over the tasks bearing each label and do not partition the \numtasks{} tasks.

\begin{table}[h]
\centering
\small
\caption{Per-label average evolved reward (\%). \emph{Left}: knowledge-type labels. \emph{Right}: task-type labels. Counts are over tasks bearing each label (multi-label).}
\label{tab:labelstats}
\begin{minipage}[t]{0.44\columnwidth}
\centering
\setlength{\tabcolsep}{4pt}
\begin{tabular}{@{}lrr@{}}
\toprule
Knowledge type & $n$ & Reward \\
\midrule
domain procedure & 58 & 69.2 \\
library api usage & 33 & 79.2 \\
mathematical method & 28 & 80.7 \\
tool workflow & 19 & 68.4 \\
file format knowledge & 17 & 82.4 \\
debugging heuristic & 8 & 70.0 \\
data cleaning procedure & 7 & 85.7 \\
evaluation protocol & 3 & 66.7 \\
\bottomrule
\end{tabular}
\end{minipage}\hfill
\begin{minipage}[t]{0.52\columnwidth}
\centering
\setlength{\tabcolsep}{4pt}
\begin{tabular}{@{}lrr@{}}
\toprule
Task type & $n$ & Reward \\
\midrule
analysis & 24 & 69.0 \\
calculation & 20 & 72.8 \\
transformation & 18 & 88.9 \\
detection & 15 & 80.0 \\
implementation & 15 & 80.0 \\
optimization & 12 & 80.0 \\
extraction & 11 & 63.6 \\
repair & 8 & 45.0 \\
planning & 8 & 57.5 \\
simulation & 7 & 71.4 \\
debugging & 6 & 76.7 \\
classification & 6 & 83.3 \\
generation & 5 & 100.0 \\
verification & 5 & 80.0 \\
search & 4 & 25.0 \\
formatting & 3 & 100.0 \\
ranking & 1 & 0.0 \\
\bottomrule
\end{tabular}
\end{minipage}
\end{table}

Among knowledge types, file format knowledge ($82.4\%$) and data cleaning procedure ($85.7\%$) attain the highest evolved rewards, while evaluation protocol ($66.7\%$, $n=3$) and tool workflow ($68.4\%$) are the lowest. Among task types, generation ($100\%$, $n=5$) and transformation ($88.9\%$) lead, whereas search ($25.0\%$, $n=4$) and repair ($45.0\%$, $n=8$) lag behind. These per-label patterns are consistent with the main-text finding (\S\ref{sec:exp:analysis}) that skills with observable execution signals benefit most from paired revision.

\paragraph{Frozen task manifest and per-task results.}
Table~\ref{tab:pertask} lists the task identifiers that define the fixed evaluation snapshot and reports the per-task rewards behind Table~\ref{tab:main} across the following pages. Dashes mark missing runs (two no-skill, one with-skill); they count as 0 in all averages (\S\ref{sec:exp:setup}). Non-integer rewards are shown to two decimal places. Domain codes: SE = Software Engineering, OWC = Office \& White Collar, NS = Natural Science, IPS = Industrial \& Physical Systems, FE = Finance \& Economics, MOR = Mathematics \& OR, CS = Cybersecurity, MCP = Media \& Content Production.

Three patterns are visible before inspecting individual rows. First, Software Engineering shows the largest domain-level improvement: evolved reward rises to $78.8\%$ from a curated-skill baseline of $40.3\%$, with six tasks flipping from $0$ to $1$. Second, Repair-mode tasks account for most reversals in which the evolved skill falls below the stronger baseline, including \texttt{crystallographic-wyckoff} (evolved $0.55$ vs.\ curated $1$) and \texttt{latex-formula-extraction} (evolved $0$ vs.\ no-skill $1$). Third, Finance \& Economics contains the highest concentration of zero-reward tasks across all three conditions, consistent with the tied domain result in Table~\ref{tab:main}.

\begingroup
\footnotesize
\setlength{\tabcolsep}{3pt}
\begin{longtable}{@{}lllccc@{}}
\caption{Per-task rewards for all \numtasks{} tasks. No skill / Curated skill / Evolved are the rewards of the corresponding condition; dashes mark missing runs, counted as 0 in all averages. Non-integer rewards are shown to two decimal places.}\label{tab:pertask}\\
\toprule
Task & Dom. & Mode & No skill & \shortstack{Curated\\skill} & Evolved \\
\midrule
\endfirsthead
\multicolumn{6}{@{}l}{\tablename~\thetable\ \textit{(continued)}} \\
\toprule
Task & Dom. & Mode & No skill & \shortstack{Curated\\skill} & Evolved \\
\midrule
\endhead
\bottomrule
\multicolumn{6}{@{}r@{}}{\textit{Continued on next page}} \\
\endfoot
\bottomrule
\endlastfoot
\texttt{azure-\allowbreak{}bgp-\allowbreak{}oscillation-\allowbreak{}route-\allowbreak{}leak} & SE & Repair & 0 & 0 & 0 \\
\texttt{data-\allowbreak{}to-\allowbreak{}d3} & SE & Repair & 0 & 0 & 1 \\
\texttt{debug-\allowbreak{}trl-\allowbreak{}grpo} & SE & Refine & 0.60 & 0.60 & 0.60 \\
\texttt{dialogue-\allowbreak{}parser} & SE & Refine & 1 & 1 & 1 \\
\texttt{fix-\allowbreak{}build-\allowbreak{}agentops} & SE & Refine & 0 & 0 & 1 \\
\texttt{fix-\allowbreak{}build-\allowbreak{}google-\allowbreak{}auto} & SE & Repair & -- & 0 & 0 \\
\texttt{fix-\allowbreak{}visual-\allowbreak{}stability} & SE & Refine & 1 & 1 & 1 \\
\texttt{flink-\allowbreak{}query} & SE & Refine & 1 & 0 & 1 \\
\texttt{jax-\allowbreak{}computing-\allowbreak{}basics} & SE & Refine & 1 & 0 & 1 \\
\texttt{llm-\allowbreak{}prefix-\allowbreak{}cache-\allowbreak{}replay} & SE & Refine & 0 & 1 & 1 \\
\texttt{parallel-\allowbreak{}tfidf-\allowbreak{}search} & SE & Refine & 1 & 1 & 1 \\
\texttt{python-\allowbreak{}scala-\allowbreak{}translation} & SE & Repair & 1 & 0 & 0 \\
\texttt{react-\allowbreak{}performance-\allowbreak{}debugging} & SE & Refine & 0 & 0 & 1 \\
\texttt{simpo-\allowbreak{}code-\allowbreak{}reproduction} & SE & Refine & 0 & 0 & 1 \\
\texttt{spring-\allowbreak{}boot-\allowbreak{}jakarta-\allowbreak{}migration} & SE & Repair & 1 & 1 & 1 \\
\texttt{tictoc-\allowbreak{}unnecessary-\allowbreak{}abort-\allowbreak{}detection} & SE & Refine & 1 & 0.85 & 1 \\
\texttt{citation-\allowbreak{}check} & OWC & Refine & 1 & 1 & 1 \\
\texttt{court-\allowbreak{}form-\allowbreak{}filling} & OWC & Repair & 1 & 0 & 1 \\
\texttt{edit-\allowbreak{}pdf} & OWC & Repair & 0 & 0 & 1 \\
\texttt{enterprise-\allowbreak{}information-\allowbreak{}search} & OWC & Repair & 0 & 0 & 0 \\
\texttt{exceltable-\allowbreak{}in-\allowbreak{}ppt} & OWC & Repair & 1 & 1 & 1 \\
\texttt{jpg-\allowbreak{}ocr-\allowbreak{}stat} & OWC & Refine & 1 & 1 & 1 \\
\texttt{latex-\allowbreak{}formula-\allowbreak{}extraction} & OWC & Repair & 1 & 0 & 0 \\
\texttt{offer-\allowbreak{}letter-\allowbreak{}generator} & OWC & Refine & 1 & 1 & 1 \\
\texttt{organize-\allowbreak{}messy-\allowbreak{}files} & OWC & Refine & 1 & 1 & 1 \\
\texttt{paper-\allowbreak{}anonymizer} & OWC & Repair & 0 & 0 & 1 \\
\texttt{pdf-\allowbreak{}excel-\allowbreak{}diff} & OWC & Repair & 1 & 1 & 1 \\
\texttt{powerlifting-\allowbreak{}coef-\allowbreak{}calc} & OWC & Refine & 1 & 1 & 1 \\
\texttt{pptx-\allowbreak{}reference-\allowbreak{}formatting} & OWC & Repair & 0 & 1 & 1 \\
\texttt{sales-\allowbreak{}pivot-\allowbreak{}analysis} & OWC & Repair & 0 & 1 & 1 \\
\texttt{taxonomy-\allowbreak{}tree-\allowbreak{}merge} & OWC & Refine & 0 & 0 & 1 \\
\texttt{crystallographic-\allowbreak{}wyckoff-\allowbreak{}position-\allowbreak{}analysis} & NS & Repair & 0.55 & 1 & 0.55 \\
\texttt{earthquake-\allowbreak{}phase-\allowbreak{}association} & NS & Refine & 1 & 1 & 1 \\
\texttt{earthquake-\allowbreak{}plate-\allowbreak{}calculation} & NS & Repair & 0 & 1 & 1 \\
\texttt{exoplanet-\allowbreak{}detection-\allowbreak{}period} & NS & Repair & 1 & 1 & 1 \\
\texttt{find-\allowbreak{}topk-\allowbreak{}similiar-\allowbreak{}chemicals} & NS & Repair & 0 & 0 & 0 \\
\texttt{flood-\allowbreak{}risk-\allowbreak{}analysis} & NS & Refine & 0 & 1 & 1 \\
\texttt{glm-\allowbreak{}lake-\allowbreak{}mendota} & NS & Repair & 1 & 1 & 1 \\
\texttt{gravitational-\allowbreak{}wave-\allowbreak{}detection} & NS & Repair & -- & -- & 1 \\
\texttt{lab-\allowbreak{}unit-\allowbreak{}harmonization} & NS & Refine & 0.69 & 1 & 1 \\
\texttt{lake-\allowbreak{}warming-\allowbreak{}attribution} & NS & Refine & 0 & 1 & 1 \\
\texttt{mars-\allowbreak{}clouds-\allowbreak{}clustering} & NS & Refine & 1 & 1 & 1 \\
\texttt{protein-\allowbreak{}expression-\allowbreak{}analysis} & NS & Refine & 0 & 1 & 1 \\
\texttt{quantum-\allowbreak{}numerical-\allowbreak{}simulation} & NS & Repair & 0 & 0 & 0 \\
\texttt{radar-\allowbreak{}vital-\allowbreak{}signs} & NS & Repair & 0 & 1 & 1 \\
\texttt{seismic-\allowbreak{}phase-\allowbreak{}picking} & NS & Refine & 1 & 0 & 1 \\
\texttt{3d-\allowbreak{}scan-\allowbreak{}calc} & IPS & Refine & 1 & 1 & 1 \\
\texttt{ada-\allowbreak{}bathroom-\allowbreak{}plan-\allowbreak{}repair} & IPS & Refine & 0 & 1 & 0 \\
\texttt{adaptive-\allowbreak{}cruise-\allowbreak{}control} & IPS & Repair & 0 & 1 & 1 \\
\texttt{drone-\allowbreak{}planning-\allowbreak{}control} & IPS & Refine & 0 & 0.57 & 0 \\
\texttt{dynamic-\allowbreak{}object-\allowbreak{}aware-\allowbreak{}egomotion} & IPS & Repair & 0 & 0 & 0 \\
\texttt{energy-\allowbreak{}ac-\allowbreak{}optimal-\allowbreak{}power-\allowbreak{}flow} & IPS & Refine & 0 & 0 & 1 \\
\texttt{energy-\allowbreak{}market-\allowbreak{}pricing} & IPS & Repair & 1 & 1 & 1 \\
\texttt{energy-\allowbreak{}unit-\allowbreak{}commitment} & IPS & Refine & 0 & 1 & 0 \\
\texttt{grid-\allowbreak{}dispatch-\allowbreak{}operator} & IPS & Refine & 0 & 1 & 1 \\
\texttt{hvac-\allowbreak{}control} & IPS & Refine & 1 & 0 & 1 \\
\texttt{manufacturing-\allowbreak{}codebook-\allowbreak{}normalization} & IPS & Repair & 0 & 1 & 1 \\
\texttt{manufacturing-\allowbreak{}equipment-\allowbreak{}maintenance} & IPS & Repair & 0 & 1 & 1 \\
\texttt{manufacturing-\allowbreak{}fjsp-\allowbreak{}optimization} & IPS & Refine & 0 & 0 & 1 \\
\texttt{r2r-\allowbreak{}mpc-\allowbreak{}control} & IPS & Repair & 0 & 0 & 1 \\
\texttt{econ-\allowbreak{}detrending-\allowbreak{}correlation} & FE & Refine & 1 & 1 & 1 \\
\texttt{financial-\allowbreak{}modeling-\allowbreak{}qa} & FE & Repair & 0 & 0 & 0 \\
\texttt{invoice-\allowbreak{}fraud-\allowbreak{}detection} & FE & Refine & 1 & 1 & 1 \\
\texttt{reserves-\allowbreak{}at-\allowbreak{}risk-\allowbreak{}calc} & FE & Repair & 0 & 0 & 0 \\
\texttt{sec-\allowbreak{}financial-\allowbreak{}report} & FE & Repair & 0 & 0 & 0 \\
\texttt{shock-\allowbreak{}analysis-\allowbreak{}demand} & FE & Repair & 0 & 0 & 0 \\
\texttt{trend-\allowbreak{}anomaly-\allowbreak{}causal-\allowbreak{}inference} & FE & Refine & 1 & 1 & 1 \\
\texttt{weighted-\allowbreak{}gdp-\allowbreak{}calc} & FE & Repair & 0 & 1 & 1 \\
\texttt{xlsx-\allowbreak{}recover-\allowbreak{}data} & FE & Repair & 0 & 0 & 0 \\
\texttt{civ6-\allowbreak{}adjacency-\allowbreak{}optimizer} & MOR & Refine & 0 & 0.85 & 0.60 \\
\texttt{exam-\allowbreak{}block-\allowbreak{}sequencing} & MOR & Refine & 0 & 0 & 1 \\
\texttt{lean4-\allowbreak{}proof} & MOR & Refine & 1 & 1 & 1 \\
\texttt{paratransit-\allowbreak{}routing} & MOR & Repair & 0 & 1 & 1 \\
\texttt{pddl-\allowbreak{}airport-\allowbreak{}planning} & MOR & Refine & 0 & 0 & 0 \\
\texttt{pddl-\allowbreak{}tpp-\allowbreak{}planning} & MOR & Repair & 1 & 0 & 1 \\
\texttt{travel-\allowbreak{}planning} & MOR & Repair & 0 & 0 & 0 \\
\texttt{virtualhome-\allowbreak{}agent-\allowbreak{}planning} & MOR & Repair & 0 & 0 & 0 \\
\texttt{dapt-\allowbreak{}intrusion-\allowbreak{}detection} & CS & Refine & 0 & 1 & 1 \\
\texttt{fix-\allowbreak{}druid-\allowbreak{}loophole-\allowbreak{}cve} & CS & Repair & 1 & 0 & 0 \\
\texttt{fix-\allowbreak{}erlang-\allowbreak{}ssh-\allowbreak{}cve} & CS & Repair & 0 & 1 & 1 \\
\texttt{setup-\allowbreak{}fuzzing-\allowbreak{}py} & CS & Repair & 0.50 & 0.83 & 0 \\
\texttt{software-\allowbreak{}dependency-\allowbreak{}audit} & CS & Refine & 0 & 0 & 1 \\
\texttt{syzkaller-\allowbreak{}ppdev-\allowbreak{}syzlang} & CS & Repair & 1 & 1 & 1 \\
\texttt{mario-\allowbreak{}coin-\allowbreak{}counting} & MCP & Refine & 0 & 1 & 1 \\
\texttt{multilingual-\allowbreak{}video-\allowbreak{}dubbing} & MCP & Repair & 1 & 1 & 1 \\
\texttt{pedestrian-\allowbreak{}traffic-\allowbreak{}counting} & MCP & Repair & 0.03 & 0.75 & 1 \\
\texttt{threejs-\allowbreak{}structure-\allowbreak{}parser} & MCP & Repair & 0 & 1 & 1 \\
\texttt{threejs-\allowbreak{}to-\allowbreak{}obj} & MCP & Refine & 1 & 1 & 1 \\
\texttt{video-\allowbreak{}silence-\allowbreak{}remover} & MCP & Repair & 0 & 0 & 0 \\
\end{longtable}
\endgroup

\medskip

\end{document}